\documentclass[10pt,twocolumn,letterpaper]{article}

\usepackage{wacv}
\usepackage{times}
\usepackage{epsfig}
\usepackage{graphicx}
\usepackage{amsmath}
\usepackage{amssymb}
\usepackage{multirow}
\usepackage{amsmath}
\usepackage{amssymb}
\usepackage{amsthm}
\usepackage{dsfont}
\usepackage{algorithm}
\usepackage{algorithmic}
\usepackage{booktabs}
\usepackage{framed}

\newenvironment{tsaligned}{%
  \begin{equation}\begin{aligned}%
}{%
  \end{aligned}\end{equation}%
}
\newenvironment{tsaligned*}{%
  \begin{equation*}\begin{aligned}%
}{%
  \end{aligned}\end{equation*}%
}
\newcommand{\tsja}[1]{}
\newcommand{\tsnote}[1]{}

  \theoremstyle{definition}
  \newtheorem{theorem}{Theorem}%% [section]
  \theoremstyle{definition}
  \newtheorem{lemma}{Lemma}% [section]
  \theoremstyle{definition}
  \newtheorem{definition}{Definition}%% % [section]
  {\endMakeFramed}

  \newenvironment{greenleftbar}{%
    \MakeFramed {\advance\hsize-\width \FrameRestore}}%
  {\endMakeFramed}

  \newenvironment{lightgrayleftbar}{%
    \MakeFramed {\advance\hsize-\width \FrameRestore}}%
  {\endMakeFramed}

  \newenvironment{example-waku}
    {\begin{lightgrayleftbar}\begin{example}}
    {\end{example}\end{lightgrayleftbar}}
  \newenvironment{exercise-waku}
    {\begin{lightgrayleftbar}\begin{exercise}}
    {\end{exercise}\end{lightgrayleftbar}}
  \newenvironment{simulation-waku}
    {\begin{greenleftbar}\begin{simulation}}
    {\end{simulation}\end{greenleftbar}}
  \newenvironment{proposition-waku}
    {\begin{oframed}\begin{proposition}}
    {\end{proposition}\end{oframed}}
  \newenvironment{definition-waku}
    {\begin{oframed}\begin{definition}}
    {\end{definition}\end{oframed}}
  \newenvironment{lemma-waku}
    {\begin{oframed}\begin{lemma}}
    {\end{lemma}\end{oframed}}
  \newenvironment{theorem-waku}
    {\begin{oframed}\begin{theorem}}
    {\end{theorem}\end{oframed}}
  \newenvironment{property-waku}
    {\begin{oframed}\begin{property}}
    {\end{property}\end{oframed}}
  \newenvironment{corollary-waku}
    {\begin{oframed}\begin{corollary}}
    {\end{corollary}\end{oframed}}
  \newenvironment{conjecture-waku}
    {\begin{oframed}\begin{conjecture}}
    {\end{conjecture}\end{oframed}}

  \newcommand{\1}{{\mathbf{1}}}
  \newcommand{\va}{{\mathbf{a}}}
  \newcommand{\vb}{{\mathbf{b}}}

  \newcommand{\vh}{{\mathbf{h}}}
  
  \newcommand{\vj}{{\mathbf{j}}}

  \renewcommand{\vs}{{\mathbf{s}}}

  \newcommand{\vt}{{\mathbf{t}}}
  \newcommand{\vu}{{\mathbf{u}}}

  \newcommand{\vv}{{\mathbf{v}}}
  \newcommand{\vw}{{\mathbf{w}}}
  \newcommand{\vwtil}{{\tilde{\mathbf{w}}}}
  \def\x{\mathbf{x}}%\newcommand{\x}{{\mathbf{x}}}% for lualatex

  \newcommand{\z}{{\mathbf{z}}}

  \newcommand{\cA}{{\mathcal{A}}}

  \newcommand{\cD}{{\mathcal{D}}}
  \newcommand{\bE}{{\mathbb{E}}}

  \newcommand{\cH}{{\mathcal{H}}}

  \newcommand{\cJ}{{\mathcal{J}}}

  \newcommand{\bN}{{\mathbb{N}}}
  \newcommand{\cN}{{\mathcal{N}}}

  \newcommand{\vP}{{\mathbf{P}}}
  \newcommand{\bR}{{\mathbb{R}}}

  \newcommand{\cS}{{\mathcal{S}}}

  \newcommand{\cT}{{\mathcal{T}}}
  \newcommand{\vT}{{\mathbf{T}}}
  
  \newcommand{\vU}{{\mathbf{U}}}
  
  \newcommand{\vV}{{\mathbf{V}}}

  \newcommand{\vW}{{\mathbf{W}}}
  
  \newcommand{\vWtil}{{\tilde{\mathbf{W}}}}

  \newcommand{\Z}{{\mathbf{Z}}}

  \newcommand{\vthet}{{\mathbf{\theta}}}

  \newcommand{\pderiv}[2]{\frac{\partial #1}{\partial #2}}

\def\diff{\mathop{}\!\mathrm{d}}
\def\ip#1{\left\langle#1\right\rangle}
\DeclareMathOperator\vect{vec}
\def\pderiv#1#2{\frac{\partial#1}{\partial#2}}
\newcommand\Var[2][]{%
  \ifx\relax\detokenize{#1}\relax%
    \mathop{\mathrm{Var}}%
  \else%
    \mathop{\mathrm{Var}}\limits_{#1}%
  \fi%
  \left[#2\right]%
}
\newcommand\Ex[2][]{%
  \ifx\relax\detokenize{#1}\relax%
    \mathop{\bE}%
  \else%
    \mathop{\bE}\limits_{#1}%
  \fi%
  \left[#2\right]%
}

\wacvfinalcopy % *** Uncomment this line for the final submission

\ifwacvfinal
\def\assignedStartPage{1} % *** Enter the assigned starting page number (instead of 9876)
\fi

\ifwacvfinal
\usepackage[breaklinks=true,bookmarks=false]{hyperref}
\else
\usepackage[pagebackref=true,breaklinks=true,colorlinks,bookmarks=false]{hyperref}
\fi

\ifwacvfinal
\else
\fi

\begin{document}

%%%%%%%%% TITLE
\title{Adaptive Signal Variances: CNN Initialization Through Modern Architectures}

\author{Takahiko Henmi, Esmeraldo Ronnie Rey Zara, Yoshihiro Hirohashi, Tsuyoshi Kato\\
  Faculty of Science and Technology, Gunma University\\
  Kiryu, Tenjin-cho 1-5-1, 376-8515, Japan\\
  {\tt\small katotsu@cs.gunma-u.ac.jp}
}

\maketitle
%\thispagestyle{empty}

%%%%%%%%% ABSTRACT
\begin{abstract}
Deep convolutional neural networks (CNN) have achieved the unwavering confidence in its performance on image processing tasks. The CNN architecture constitutes a variety of different types of layers including the convolution layer and the max-pooling layer. CNN practitioners widely understand the fact that the stability of learning depends on how to initialize the model parameters in each layer. Nowadays, no one doubts that the de facto standard scheme for initialization is the so-called Kaiming initialization that has been developed by He et al. The Kaiming scheme was derived from a much simpler model than the currently used CNN structure having evolved since the emergence of the Kaiming scheme. The Kaiming model consists only of the convolution and fully connected layers, ignoring the max-pooling layer and the global average pooling layer. In this study, we derived the initialization scheme again not from the simplified Kaiming model, but precisely from the modern CNN architectures, and empirically investigated how the new initialization methods perform compared to the de facto standard ones that are widely used today. 
\end{abstract}

%%%%%%%%% BODY TEXT
\section{Introduction}

Deep convolutional neural networks (CNN) have established an unshakable confidence in its power for computer vision including image classification~\cite{Kleinberg-icml18a,Szegedy-cvpr15a,Simonyan-iclr15a}, object detection~\cite{Girshick-iccv15a,ItoSatSan18}, segmentation~\cite{Kaiming-cvpr16a}, and retrieval tasks~\cite{ShengyongDing-patcog15a}. In the past decade, its impressive performance has attracted attention not only of many academic researchers but also a huge number of industrial companies dominating the world. These days it is hard to find a top racer who does not exploit a deep CNN in world-wide computer vision competitions~\cite{Russakovsky-ilsvrc15}.

The predecessor of the current deep CNN structures is the so-called LeNet that is the first neural network developed by LeCun et al~\cite{LeCun-nips89a}. LeNet constitutes the convolution layer and fully connected layer. The hyperbolic tangent function~\cite{LeCunBOM12} is used for activation. LeNet bears a resemblance to primate's visual cortex~\cite{Hubel68a}. The current deep CNN structures made more artificial compared to LeNet. The activation function is now replaced by ReLU~\cite{NairHinton-icml10} in the modern structure. The fully connected layer is often replaced by the global average pooling~\cite{NairHinton-icml10} in order to suppress the overfitting caused by the excessive flexibility of the fully connected layer. 
The power of the deep CNN has been enhanced by increasing the capacity of the model, although the deeper structure often sacrifices the numerical stability of optimization. Compared to the hyperbolic tangent or the sigmoid activation functions, the use of ReLU activation may more likely avoid the vanishing gradient problem~\cite{Hochreiter-iju98,NairHinton-icml10}. Compared to the batch gradient descent, the stochastic gradient descent may avoid bad local minima and saddle points with help of its perturbation effect~\cite{Kleinberg-icml18a}.  However, these two techniques are not good enough to achieve a stable convergence.

Due to the high non-convexity of the empirical risk function, how to initialize the model parameters including weights and biases affects the stability of optimization. In AlexNet, all the weight parameters are initialized using the zero-mean normal distribution with a constant variance parameter $0.1^{2}$, and all the initial bias parameters are set to one \cite{Krizhevsky2012a}. Glorot et al~\cite{Glorot-icai10a} proposed an initialization method adaptive to the network structure. Their method is referred to as the Xavier method. They regard the network parameters as random variables to derive their initialization method. The result of their analysis is that the inverse of the variance parameter for the zero-mean normal should be the average of the numbers of input units and output units for each layer. In Glorot's model, the hyperbolic tangent function is assumed to be used for activation in every layer, which does not match the fact that the activation function has been replaced to ReLU or its variants and the hyperbolic tangent and the sigmoid activation have been disposed in most of the currently used CNN structures.

At present, the \textit{de facto} standard initialization method is the Kaiming method that was proposed by He et al~\cite{Kaiming-iccv15}. They substituted the ReLU function to the activation function and analyzed the random network in the fashion similar to Glorot et al's analysis. Kaiming method enables us to train deeper structures stably which cannot be optimized by Xavier method. However, Kaiming method is still much simplified compared to the greatly evolved current CNN model. Their model does not account for the subsampling layers such as the max pooling layer and the global average pooling layer, which gives rise to two questions; (i) Is it possible to derive an initialization method from a more precise CNN model? (ii) How does the gradient descent work when using a more precise initialization model? 

In this paper, we propose a new initialization method which takes account the subsampling layers, whether it is the max pooling or the global average pooling layer. We show that this proposed initialization method adopts a more precise CNN structure that analyzes the signal variances for the forward and backward signals. We show that this proposed method is an extension of the Kaiming method, includes all its basic components and takes into account the padding operation.
We empirically investigate how the new initialization method performs compared to the de facto standard ones that are widely used today.

\section{Related Work}
The most popular initialization methods are the Xavier method and the Kaiming method. Both of them are derived from random networks drawn from normal distributions. Similar random models have long been used to understand optimization and generalization of deep neural structures so far. 

The difficulty of the analysis of neural networks lie in the non-convex empirical risks. The alternative is the convex formulation of the empirical risk that is the network without hidden layers. Conventional linear models such as the logistic regression, the support vector machine and their kernelized models belong to the network without hidden layers. All local minima of such networks are global minima, which eliminates the dependency on the initial solution~\cite{Venturi-jmlr19,YuanzhiLi-arxiv17}. In contrast, deep structures suffer from many local minima, which means that starting the gradient descent from different initial guesses yields different solutions. 

The reason why many theoretical works employed the normal distributions is from its strong connection with Gaussian process~\cite{Neal-book12,JaehoonLee-iclr18a,Koriyama-ast20a,Blomqvist-mlkdd20}. Some researches considered over-parameterized models in which all the local minima approach to zero meaning that they all are nearly a global minimum~\cite{SimonDu-nips18,DifanZou-arxiv18a}.
%% [Du et al., 2019, 2018b, Du and Hu, 2019, Li and Liang, 2018, Allen-Zhu et al., 2018b,a, Zou et al., 2018, Cao and Gu, 2019].
Another advantage of over-parameterization is that the gradient descent rapidly converges to a local minimum because any of the approximately initialized solutions are near to the local minimum in such a model~\cite{DifanZou-arxiv18a}. However, such a model consumes a large space complexity, which requires expensive computational resources for choosing an over-parameterized CNN structure. Some theoretical studies attempt to understand the global landscape of the empirical risk function~\cite{LiangShiyu-icm18a}. They assume a single hidden layer, although many reports empirically showed that deeper structures perform better than shallow structure for computer vision~\cite{Kaiming-cvpr16a}.

Thus, the CNN structures practically trainable with a reasonable computational resource differ far away from the models assumed in theoretical works. In contrast, our analysis yields a practical method, yet the assumption used in this study is realistic. 
\begin{figure*}
  \centering
  \begin{tabular}{ll}
    (a) Convolution & (b) Fully Connected
    \\
    \includegraphics[width=.45\linewidth]{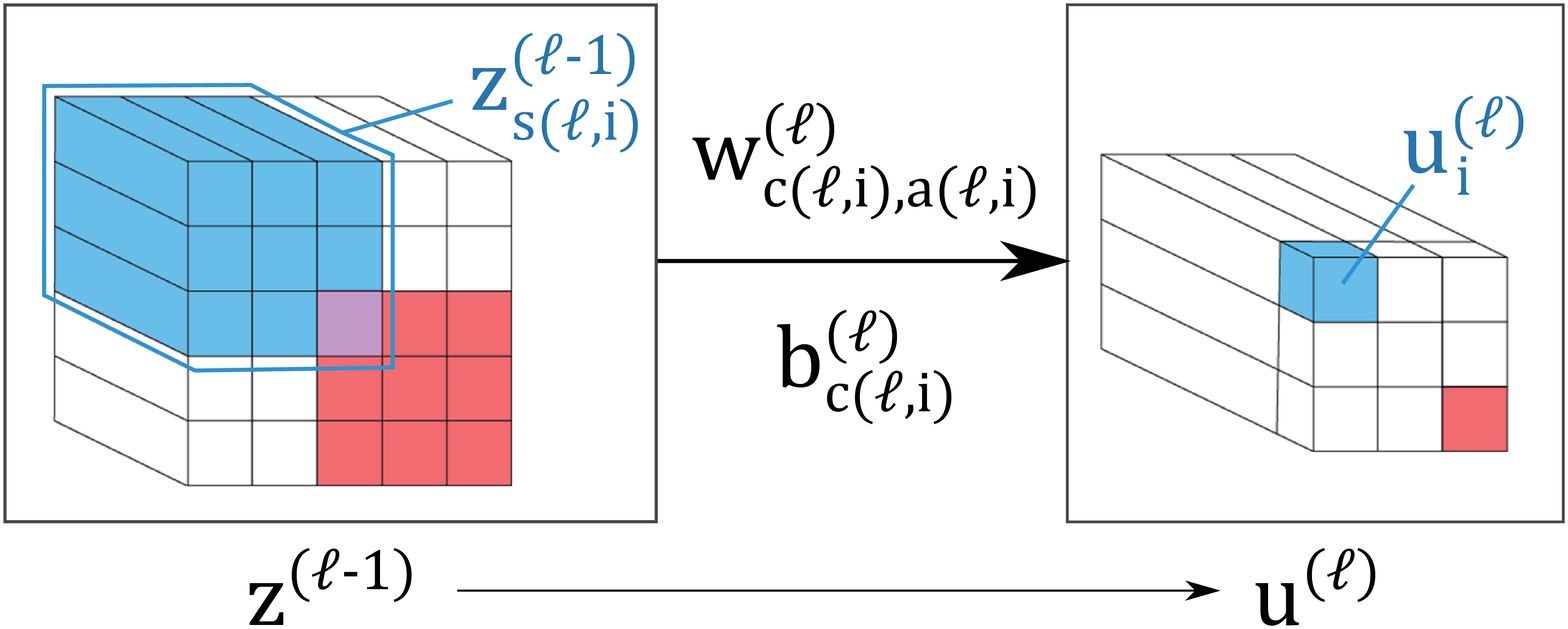}
    &
    \includegraphics[width=.45\linewidth]{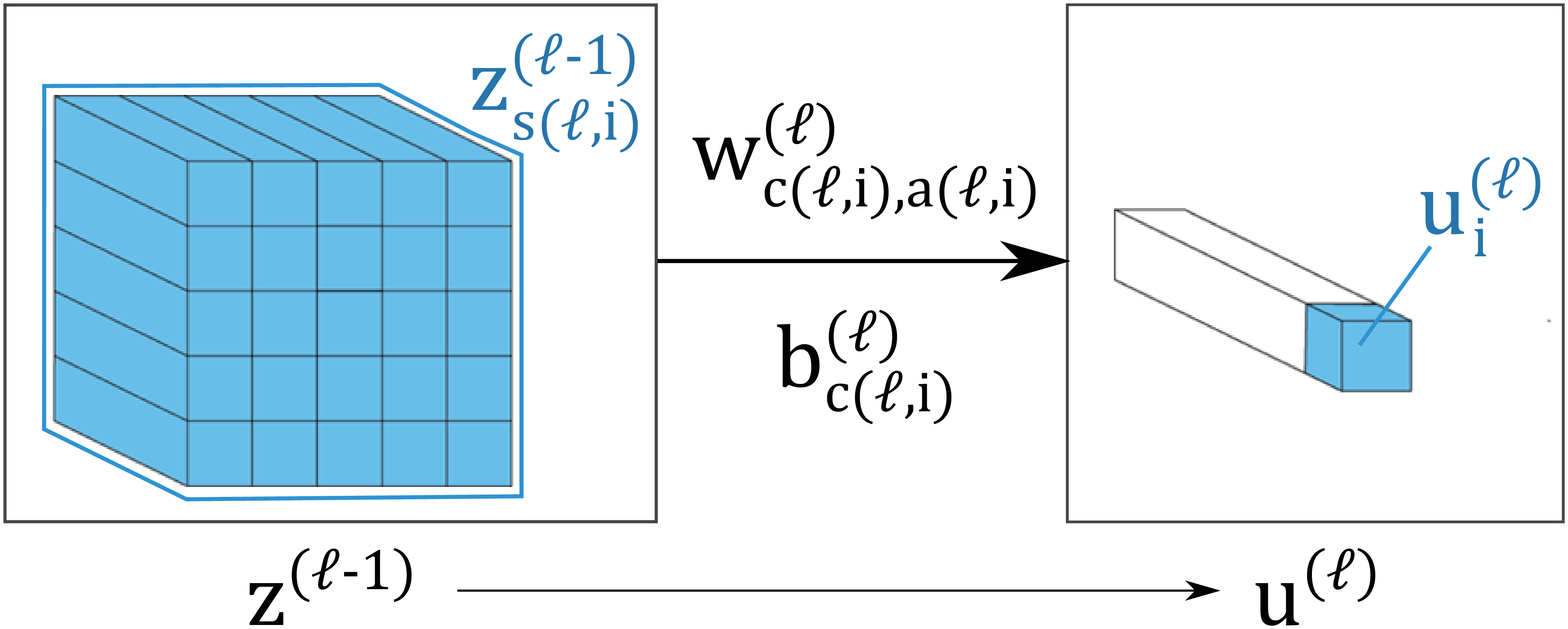}
    \\ 
    \\
    (c) Activation function & (d) Pooling operations
    \\
    \includegraphics[width=.45\linewidth]{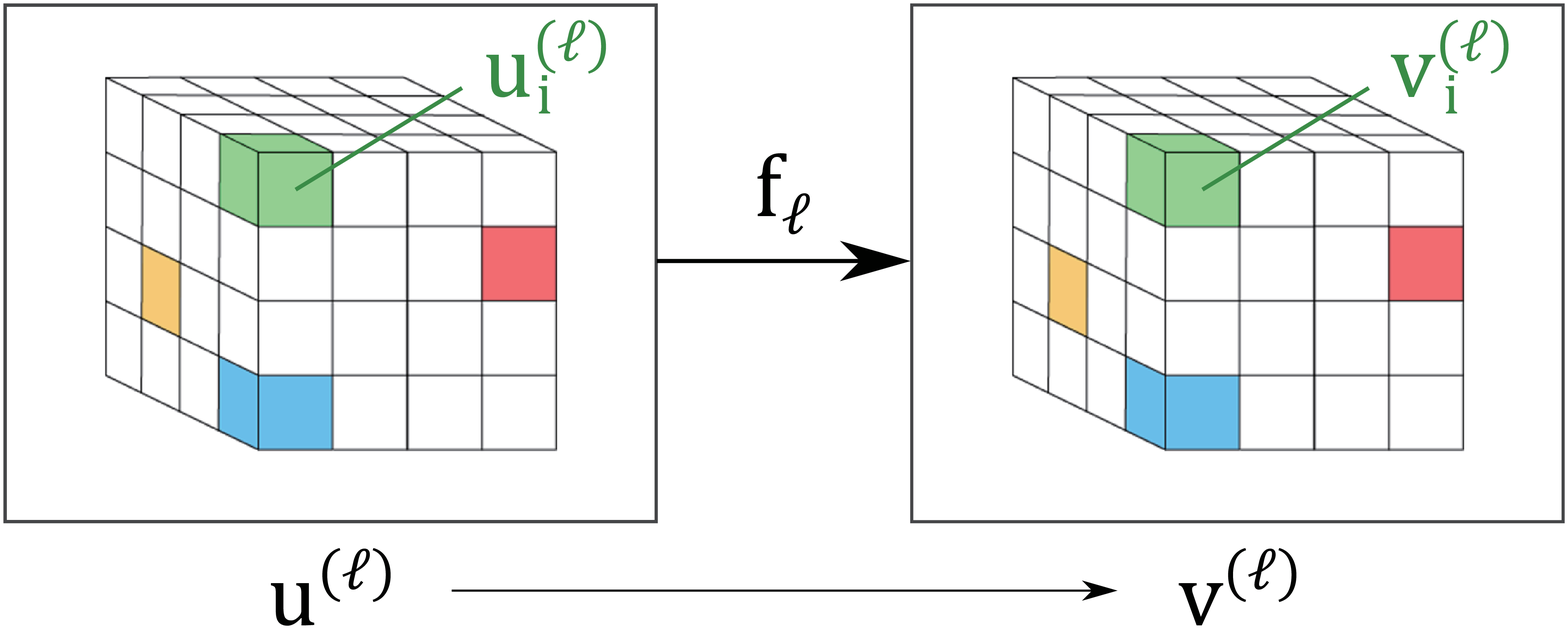}
    &
    \includegraphics[width=.45\linewidth]{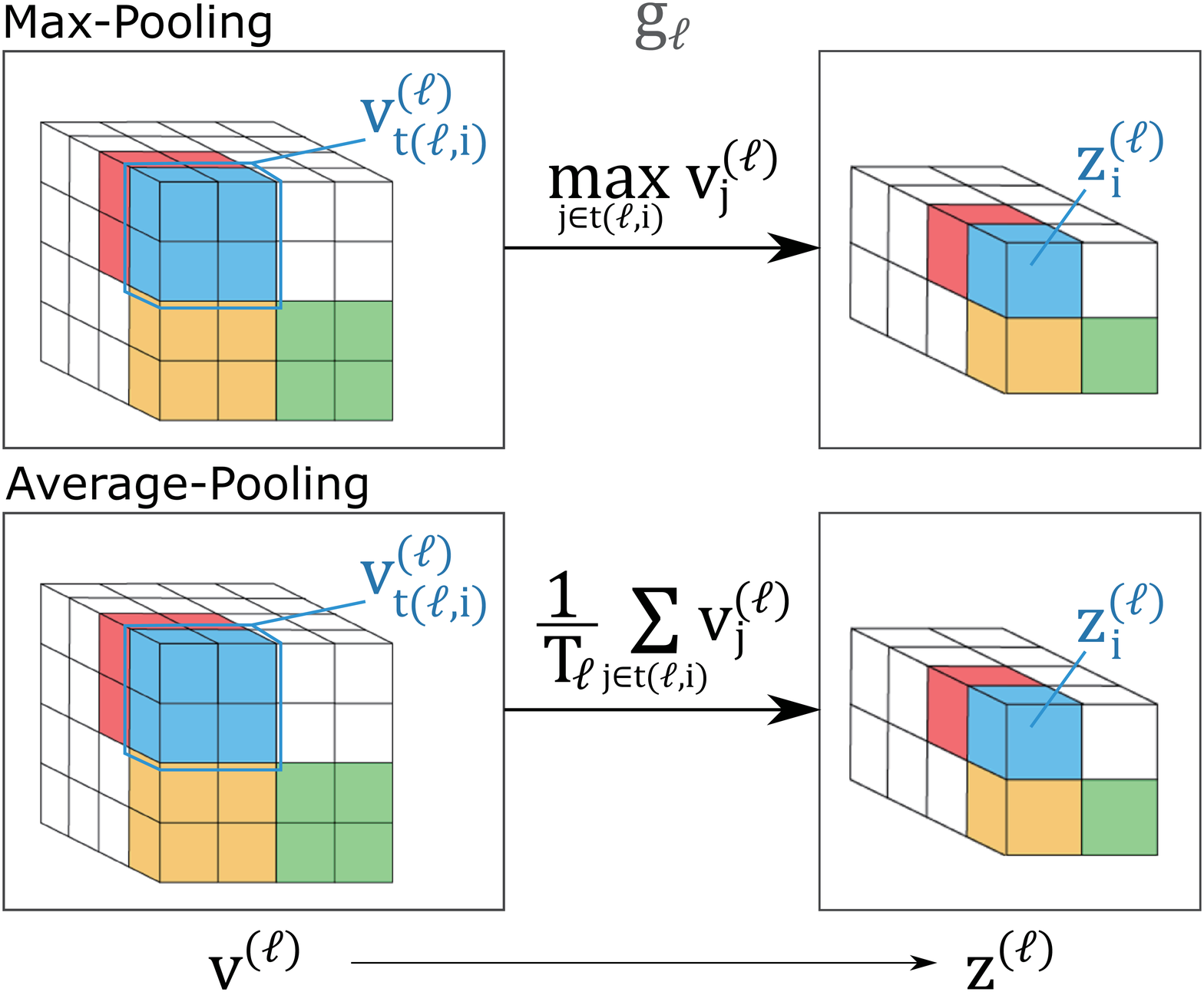}
  \end{tabular}
  \caption{Our CNN model.
    Panel (a) depicts a convolution operation that computes
    a unit $u_{i}^{(\ell)}$ in $\ell$th layer.
    The $i$th unit is in $c(i,\ell)$th channel.
    The convolution takes a part of signals in $\z^{(\ell-1)}\in M_{\ell}$. 
    The part is specified by a index set $\vs(\ell,i)$.
    Panel (b) depicts a fully connected layer that is 
    a special case of the convolution operation
    where $\vs(\ell,i)=[M_{\ell}]$.
    Panel (c) explains an activation function $f_{\ell}$
    such as sigmoid and ReLU. Activation functions
    transform signals component-wise.
    In the panel (d), the max pooling and the average
    pooling, denoted by $g_{\ell}$, are explained.
    Pooling functions aggregates a part of signals $\vv^{(\ell)}\in M'_{\ell}$.
    The sub-vector is specified by the index set $\vt(\ell,i)$.
    \label{fig:formulation-visualized}}
\end{figure*}
\begin{table*}[t!]
  \caption{Notation used for describing the forward propagation. 
    \label{tab:fornot} }
  \centering
  \begin{tabular}{rl}
    \\
    \hline
    Symbol & Description
    \\
    \hline \hline
  $C_{\ell} \in\bN$ & Number of channels.
  \\
  $S_{\ell} \in\bN$ & Receptive field size.
  \\
  %% $\vw^{(\ell)}_{i} \in \bR^{S_{\ell}}$ & Convolutional kernel ($i=1,\dots,C_\ell$). 
  %% \\
  %% $\vb^{(\ell)}\in\bR^{C_{\ell}}$ & Bias. 
  %% \\
  $c(\ell,i)\in[C_\ell]$         & Channel containing $i$th unit. 
  \\
  $\va(\ell,i)\subseteq[S_\ell]$ & Index set for convolution kernel. 
  \\
  $\vs(\ell,i)\subseteq[M_{\ell-1}]$ & Index set of units fed to the convolution operation
  producing $u_{i}^{(\ell)}$. $|\vs(\ell,i)|=|\va(\ell,i)|$. 
  \\
  $M^\prime_\ell\in\bN$ & Number of units just after convolution. 
  \\
  %% $\vu^{(\ell)}\in\bR^{M^\prime_\ell}$ & Signals just after convolution. 
  %% \\
  $f_\ell:\bR\to\bR$ & Activation function.
  \\
  %% $\vv^{(\ell)}\in\bR^{M^{\prime}_{\ell}}$ & Activation. 
  %% \\
  $T_\ell\in\bN$ &  Size of aggregation region.
  \\
  $g_\ell:\bR^{T_{\ell}}\to\bR$ & Pooling function. 
  \\
  $\vt(\ell,i)\subseteq[M^\prime_\ell]$ & Aggregation region. 
    The subsets divides $[M^\prime_\ell]$ exclusively. $|\vt(\ell,i)|=T_{\ell}$.
  \\
  %% $\z^{(\ell)}\in\bR^{M_\ell}$ & Signals transformed in $\ell$th layer. 
  %% \\
  $M_{\ell}\in\bN$ & Size of signals $\z^{(\ell)}$.
  \\
  \hline
\end{tabular}
\end{table*}
\begin{table*}[t!]
  \caption{Notation used for describing the backward propagation. 
    \label{tab:backnot} }
  \centering
  \begin{tabular}{rl}
    \\
    \hline
    Symbol & Description
    \\
    \hline \hline
    %% $\Delta\z^{(\ell)}=\frac{\partial E}{\partial \z^{(\ell)}}$ & Gradient wrt $\z^{(\ell)}$. 
    %% \\
    %% $\Delta\vu^{(\ell)}=\frac{\partial E}{\partial \vu^{(\ell)}}$ & Gradient wrt $\vu^{(\ell)}$.
    %% \\
    %% $\Delta\vv^{(\ell)}=\frac{\partial E}{\partial \vv^{(\ell)}}$ & Gradient wrt $\vv^{(\ell)}$.
    %% \\
    $J_\ell\in\bN$ & Size of backward weights for convolution. 
    \\
    $\vwtil_{c}^{(\ell)}\in\bR^{J_\ell}$ & Backward weights for convolution.
    % The backward weights are a permutation of entries in $\vw^{(\ell)}$. 
    \\
    $\tilde{C}_{\ell}\in\bN$ & Number of backward weights for convolution. 
    \\
    $\tilde{c}(\ell,i)\in[\tilde{C}_\ell]$ & Channel index of backward weights. 
    \\
    $\vh(\ell,i)\subseteq[J_\ell]$ & Index set of backward weights for convolution. 
    \\
    $\vj(\ell,i)\subseteq[M^\prime_\ell]$ & Index set of backward signals for convolution.
    $|\vj(\ell,i)|=|\vh(\ell,i)|$. 
    \\
    $d(\ell,i)\in[M_\ell]$ & Unit index propagating a backward signal to $i$th unit in pooling.
    \\
  \hline
\end{tabular}
\end{table*}
\newcommand{\tschecked}{\includegraphics[width=1em]{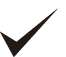}}
\newcommand{\tscheckedshiro}{\includegraphics[width=1em]{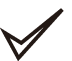}}

\begin{table}[t!]
  \centering
  \caption{%
    Comparison with the existing methods. %
    The models deriving the existing methods do not include some components of CNN, although the model in this study expresses the CNN more precisely by fully containing all components of CNN. FC and Conv are the abbreviations of the fully connected layer and the convolution layer, respectively. The Kaiming model contains the convolution layer, although the padding operation is not considered.  \label{tbl:compare-init}} 
  \begin{center}
    \begin{tabular}{|c|c|c|c|c|}
      \hline
      Methods  & FC & ReLU & Conv & Pooling \\
      \hline
      Xavier   & \tschecked & & &  \\
      \hline
      Kaiming  & \tschecked & \tschecked & \tscheckedshiro & \\
      \hline
      Proposed & \tschecked & \tschecked & \tschecked & \tschecked \\
      \hline
    \end{tabular}
  \end{center}
\end{table}
\section{CNN Models}\label{sec:cnnmdl}
In this section, we re-formulate CNN structures to clarify the range of the CNN extensions we account for in this study. Usually, CNN is expressed as a chain of many different types of components such as the convolution layer, the activation unit, the pooling layer, and the fully connected layer. In our formulation, every layer, named \emph{Conv+Pool} layer is homogeneous and capable to perform convolution, activation, and pooling. As seen later, the Conv+Pool layer can express the convolution layer without the pooling operation, the global average layer as well as the fully connected layer. Hereinafter, we may refer to the Conv+Pool layer as a layer. 

When using CNN for image processing tasks, two-dimensional convolution is performed. For speech recognition, the number of dimensions of the convolution operation is one \cite{Kalchbrenner-acl14}. In general, input signals and intermediate signals traversing in CNN are tensors. The order of tensors expressing signals depends on the applications. To make the notation simpler and to unify the formulations for any order of tensors, all signals are unfolded into a vector in our formulation. 

\subsection{Forward Signals}
Let $L$ be the number of the Conv+Pool layers in a CNN structure.
The $\ell$th layer has weight parameters
$\vW^{(\ell)}:=\left[\vw_{1}^{(\ell)},\dots,\vw_{C_{\ell}}^{(\ell)}\right]^\top\in\bR^{C_{\ell}\times S_{\ell}}$ 
and bias parameters
$\vb^{(\ell)}:=\left[b_{1}^{(\ell)},\dots,b_{C_{\ell}}^{(\ell)}\right]^\top\in\bR^{C_{\ell}}$. 
The $\ell$th Conv+Pool layer transforms the signal $\z^{(\ell-1)}\in\bR^{M_{\ell-1}}$ fed to the layer, and conveys the transformed signal to the next layer.
The input signals of the entire structure $\z^{(0)}$ are first fed to the first layer
and finally signals $\z^{(L)}\in\bR^{M'_{L}}$ are transmitted as
shown in Figure~\ref{fig:forbacksig}(a). 
\begin{figure*}
(a) Forward signals
\begin{tsaligned}
\z^{(0)}\in\bR^{M_{0}}
& 
\stackrel{\text{Conv}}{\mapsto} 
\vu^{(1)}\in\bR^{M'_{1}} 
\stackrel{\text{Act}}{\mapsto} 
\vv^{(1)}\in\bR^{M'_{1}} 
\stackrel{\text{Pool}}{\mapsto} 
\z^{(1)}\in\bR^{M_{1}} 
\stackrel{\text{Conv}}{\mapsto} \cdots
\\
& \stackrel{\text{Conv}}{\mapsto} 
\vu^{(L-1)}\in\bR^{M'_{L-1}} 
\stackrel{\text{Act}}{\mapsto} 
\vv^{(L-1)}\in\bR^{M'_{L-1}} 
\stackrel{\text{Pool}}{\mapsto} 
\z^{(L-1)}\in\bR^{M_{L-1}} 
\stackrel{\text{Conv}}{\mapsto} 
\vu^{(L)}\in\bR^{M'_{L}}. 
\end{tsaligned}
(b) Backward signals. 
\begin{tsaligned}
\Delta \vu^{(L)}\in\bR^{M'_{L}}
&
\stackrel{\text{}}{\mapsto} 
\Delta \z^{(L-1)}\in\bR^{M_{L-1}}
\stackrel{\text{}}{\mapsto} 
\Delta \vv^{(L-1)}\in\bR^{M'_{L-1}} 
\stackrel{\text{}}{\mapsto} 
\dots
\\
&
\stackrel{\text{}}{\mapsto} 
\Delta \z^{(1)}\in\bR^{M_{1}}
\stackrel{\text{}}{\mapsto} 
\Delta \vv^{(1)}\in\bR^{M'_{1}} 
\stackrel{\text{}}{\mapsto} 
\Delta \vu^{(1)}\in\bR^{M'_{1}}. 
\end{tsaligned}
\caption{Forward signals and backward signals. The first forward signal $\z^{(0)}$ is the input of CNN such as an RGB image. In that case, $M_{0}$ is three times the number of pixels in the RGB image. For the application of multi-category classification, $M^{\prime}_{L}$. \label{fig:forbacksig}}
\end{figure*}
For convenience of notation,
we add two signal vectors in the last layer,   
$\vv^{(L)}:=\vu^{(L)}$ and $\z^{(L)}:=\vv^{(L)}$. 
The following equations \eqref{eq:def:layer-conv},
\eqref{eq:def:layer-act}, and
\eqref{eq:def:layer-pool}, respectively,
express the operations of convolution,
activation, and pooling: 
\begin{align}
  \label{eq:def:layer-conv}
    u^{(\ell)}_i&=\left<\vw^{(\ell)}_{c(\ell,i),\va(\ell,i)},
      \z^{(\ell-1)}_{\vs(\ell,i)}\right>+b^{(\ell)}_{c(\ell,i)},
      \\
  \label{eq:def:layer-act}      
    v^{(\ell)}_i&=f_\ell\left(u^{(\ell)}_i\right),
      \\
  \label{eq:def:layer-pool}
    z^{(\ell)}_{i}&=g_\ell\left(\vv^{(\ell)}_{\vt(\ell,i)}\right).
\end{align}
where we have used the notation described in Table~\ref{tab:fornot}. 
Therein,
$\vw^{(\ell)}_{c(\ell,i),\va(\ell,i)}$ is a subvector of
$\vw_{c(\ell,i)}^{(\ell)}\in\bR^{S_{\ell}}$ with its entries
specified by a set of indices $\va(\ell,i)$;
the symbol $z^{(\ell)}_{i}$ is the $i$th entry of the transformed signal $\z^{(\ell)}$; 
the symbol $\z^{(\ell-1)}_{\vs(\ell,i)}$ is a subvector of the signal $\z^{(\ell-1)}$
with entries corresponding to a set of indices~$\vs(\ell,i)\subseteq [S_{\ell}]$.
Each operation is detailed in Section~\ref{sec:detailed-formulation}. 
Here, brief descriptions are given. 
\begin{itemize}
\item \textbf{Convolution: }
In case that the previous layer has a single channel, 
the convolution operation
for obtaining the signals sent to the channel $c$ in the next layer
can be expressed as the inner-product between a kernel
$\vw_{c}^{(\ell)}\in\bR^{S_{\ell}}$ and a subvector of the fed signal $\z^{(\ell-1)}$.
In case that the previous layer has $C_{\ell-1}$ channels,
the vector $\vw_{c}^{(\ell)}\in\bR^{S_{\ell}}$ is the concatenation
of $C_{\ell-1}$ kernels. 
For the majority entries in $\vu^{(\ell)}\in\bR^{}$.
the index set for the kernel in \eqref{eq:def:layer-conv}
is set to $\va(\ell,i)=[S_{\ell}]$.
For representing the \emph{padding}, 
the index set is setting to a subset $\va(\ell,i)\subsetneq[S_{\ell}]$.
See Figure~\ref{fig:formulation-visualized}(a). 
\item \textbf{Activation: }
A typical choice for the function $f_{\ell}:\bR\to\bR$ is ReLU.
Function $f_{\ell}$ is the identity function for the layer without activation.
See Figure~\ref{fig:formulation-visualized}(c). 
\item \textbf{Pooling: }
  Function $g_{\ell}:\bR\to\bR$ performs subsampling by aggregating a part of entries in the nonlinearly transformed signals $\vu^{(\ell)}\in\bR^{M'_{\ell}}$. The subsampling part is specified by the index set $\vt(\ell,i)\subset[M'_{\ell}]$.
See Figure~\ref{fig:formulation-visualized}(d).   
\end{itemize}
The fully connected layer is a special case of the Conv+Pool layer,
which can be shown by letting $\va(\ell,i):=[M_{\ell}]$ and $\vs(\ell,i):=[M_{\ell}]$ 
and by setting $f_{\ell}$ and $g_{\ell}$ to the identity function.
See Figure~\ref{fig:formulation-visualized}(b).

\subsection{Backward Signals}
CNN is trained by minimizing the empirical risk $E$ with respect to the model parameters $\vthet=(\vW^{(\ell)},\vb^{(\ell)})_{\ell\in[L]}$. The stochastic gradient descent applied to neural networks is called the back-propagation algorithm that conveys the signals $\Delta\z_{i}^{(\ell)}$ in the backward direction, as
shown in Figure~\ref{fig:forbacksig}(b), where we have defined
the backward signals as
\begin{tsaligned}
  \Delta\z^{(\ell-1)}=\pderiv{E}{\z^{(\ell-1)}},
  ~
  \Delta\vu^{(\ell)}=\pderiv{E}{\vu^{(\ell)}},
  ~
  \Delta\vv^{(\ell)}=\pderiv{E}{\vv^{(\ell)}}, 
\end{tsaligned}
The backward signals are conveyed with the following recursive expressions:
\begin{align}
  \label{eq:formulation:conv-backward}  
    \Delta z^{(\ell-1)}_i&=\left<
      \vwtil^{(\ell)}_{\tilde{c}(\ell,i),\vh(\ell,i)},
      \Delta\vu^{(\ell)}_{\vj(\ell,i)}\right>,\\
    \Delta u^{(\ell)}_i&=\Delta v^{(\ell)}_i
      \pderiv{v^{(\ell)}_i}{u^{(\ell)}_i},\\
    \Delta v^{(\ell)}_i&=\Delta z^{(\ell)}_{d(\ell,i)}
      \pderiv{z^{(\ell)}_{d(\ell,i)}}{v^{(\ell)}_i}, 
\end{align}
where we have used the notation described in Table~\ref{tab:backnot}. In Section ~\ref{sec:detailed-formulation}, we describe the details on how the back propagation rules are derived from the standard tensor representation of the CNN structure. 
\section{Initialization Methods}
In this section, we present two new methods for initializing the model parameters $\vthet=(\vW^{(\ell)},\vb^{(\ell)})_{\ell\in[L]}$. Following the Xavier and Kaiming methods, the initial values are set to random numbers generated with zero-mean normal distributions:
\begin{tsaligned}\label{eq:init-probmdl}
  w^{(\ell)}_{i,j}\sim\cN(0,\sigma^2_{w^{(\ell)}}),\quad
  b^{(\ell)}_i\sim\cN(0,\sigma^2_{b^{(\ell)}}).  
\end{tsaligned}
Notice that the variance parameters of the weight parameters and the bias parameters in the same layer are common.
The existing methods assume that the model parameters approximately follow the distribution~\eqref{eq:init-probmdl} during the gradient descent. 
% This approximation is justified if a local minimum is near to the initial solution. 

\subsection{Variances of Signals}
\label{ss:var-sig}
Under the assumption~\eqref{eq:init-probmdl}, the variances of the forward and backward signals can be analyzed. In order to avoid signal vanishing or signal explosion, the existing methods determine the two initialization parameters, $\sigma^2_{w^{(\ell)}}$ and $\sigma^2_{b^{(\ell)}}$ so that the variances of the forward or backward signals are kept approximately equal. Along this line, we developed an initialization method. The difference from the existing methods is that a more precise CNN structure is adopted to analyze the signal variances. 
Below we present recursive expressions of the signal variances and then the new initialization methods. The details of the derivations are given in Section~\ref{sec:variance}. 

\textbf{Variances of forward signals: }
Let us denote the variance of, $u_{i}^{(\ell)}$, $i$th unit in the $\ell$th layer by $q^{(\ell)}_{i}$. Suppose that, for the previous layer (i.e. $(\ell-1)$th layer), each signal in $\vu^{(\ell-1)}$ follows independently according to an identical distribution $\cN(0,q^{(\ell-1)})$ where $q^{(\ell-1)}$ is the average of $M'_{\ell-1}$ quantities $q^{(\ell-1)}_{1},\dots, q^{(\ell-1)}_{M'_{\ell}}$. If assuming that $q^{(\ell)}_{i}= q^{(\ell)}$ for all units in $\ell$th layer, we have that:
\begin{tsaligned}\label{eq:var-forward}
  q^{(\ell)}=\sigma^2_{b^{(\ell)}}
    +\sigma^2_{w^{(\ell)}}q^{(\ell-1)}\tau_{\ell-1}
    \frac{1}{M^\prime_\ell}\varepsilon_\ell, 
\end{tsaligned}
where $\tau_{\ell}$ is a constant defined as
\begin{tsaligned}
  \tau_{\ell}
  :=
    T_\ell\int_0^\infty
    s^2\phi(s)\Phi(s)^{T_\ell-1} d s
\end{tsaligned} 
if $g_{\ell}$ is the max pooling function, and 
\begin{tsaligned}
  \tau_{\ell}
  :=
    \frac{1}{2T_\ell}
    \left(1+\frac{T_\ell-1}{\pi}\right)
\end{tsaligned}
if $g_{\ell}$ is the average pooling function. 
Therein, $T_\ell$ is the size of the pooling range;  
$\phi,\Phi$ is the normal density function and its cumulative; 
the definition of $\varepsilon_{\ell}$ is the total number of connections between the $(\ell-1)$th and $\ell$th layers. 

Let us verify how appropriate the assumptions and the approximations used to derive \eqref{eq:var-forward}. 
The quantities $q^{(\ell)}_{1},\dots, q^{(\ell)}_{M'_{\ell}}$ are not necessarily be equal to each other (i.e. $q^{(\ell)}_{i}\ne q^{(\ell)}_{j}$ for $i\ne j$), although most of $q^{(\ell)}_{i}$ are equal to each other when the image size is enough large.  The normality assumption is justified if channels in the layer are many.

\textbf{Variances of backward signals: }
Assume that each signal in $\Delta\z^{(\ell)}$ is identically and independently drawn according to $\cN(0,r^{(\ell)})$ and that the statistical independence between $\Delta z^{(\ell)}_{d(\ell,j)}$ and $\partial z^{(\ell)}_{d(\ell,j)}/\partial u^{(\ell)}_j$ for $j\in\vj(\ell,j)$. Under this assumption, we obtain that 
\begin{tsaligned}\label{eq:var-backward}
  r^{(\ell-1)}=\sigma^2_{w^{(\ell)}}r^{(\ell)}
    \gamma_\ell\frac{1}{M_{\ell-1}}\varepsilon_\ell
\end{tsaligned}
where $\gamma_{\ell}$ is defined as 
\begin{tsaligned}
  \gamma_{\ell}
  :=
  \frac{2^{T_\ell}-1}{T_\ell2^{T_\ell}}
\end{tsaligned}    
if $g_{\ell}$ is the max pooling function, and 
\begin{tsaligned}
  \gamma_{\ell}
  :=
  \frac{1}{2T_\ell^2}
\end{tsaligned}
if $g_{\ell}$ is the average pooling function. 

\subsection{Proposed Initialization Methods}
\label{ss:proposed-meth}
The above observations lead to the methods for maintaining the variances of the signals. Here, we choose the initial bias parameters to $\vb^{(\ell)}=0$, corresponding to $\sigma_{b^{(\ell)}}^{2}=0$, and $q^{(0)}=r^{(L)}=1$. With these additional conditions, it can be seen that the two following methods, respectively, keep $q^{(\ell)}=1$ and $r^{(\ell)}=1$ for all layers. 

\begin{oframed}
  \textbf{ASV Forward Method:}
  \begin{equation}
    \label{eq:init-forward}
    \sigma^2_{w^{(\ell)}}=\frac{M^\prime_\ell}{\tau_{\ell-1}\varepsilon_\ell}.
  \end{equation}
\end{oframed}
  
\begin{oframed}
  \textbf{ASV Backward Method:}
  \begin{equation}
    \label{eq:init-backward}
    \sigma^2_{w^{(\ell)}}=\frac{M_{\ell-1}}{\gamma_\ell\varepsilon_\ell}.
  \end{equation}
\end{oframed}

\begin{table*}[t!]
  \centering
  \caption{
    Structure of 34-layer architecture. In this model, Max Pooling is used in layer $\ell = 1$, Global Average Pooling is used in the last layer of the feature extraction $\ell = 33$ and stride is used in the other convolution layers to change the resolution of feature map. 
    \label{tab:arch}}
  \begin{tabular}{|c|}
    \hline
    Component \verb|InputBlock|($c$)\\
    \hline
    Conv(7x7,channels=$c$,padding=3,stride=2),\\
    MaxPool(3x3,padding=1,stride=2)\\
    \hline
  \end{tabular}\\\vspace{1ex}
  \begin{tabular}{|c|}
    \hline
    Component \verb|ConvBlock|($c,s=1$)\\
    \hline
    Conv(3x3,channels=$c$,padding=1,stride=$s$),\\
    Conv(3x3,channels=$c$,padding=1,stride=1)\\
    \hline
  \end{tabular}\\\vspace{1ex}
  \begin{tabular}{|c|c|l|}
    \hline
    $\ell$-th Layer&Output Shape&34-layer Architecture\\
    \hline
    \hline
    &(3,224,224) &Input Image\\
    \hline
    1&(64,112,112)&\verb|InputBlock|($c=64$)\\
    \hline
    $2$--$7$&(64,56,56)  &\verb|ConvBlock|($c=64$)$\times3$\\
    \hline
    $8$--$15$&(128,28,28)&\verb|ConvBlock|($c=128,s=2$)\\
    &&\verb|ConvBlock|($c=128$)$\times3$\\
    \hline
    $16$--$27$&(256,14,14)&\verb|ConvBlock|($c=256,s=2$)\\
    &&\verb|ConvBlock|($c=256$)$\times5$\\
    \hline
    $28$--$33$&(512,7,7)&\verb|ConvBlock|($c=512,s=2$)\\
    &&\verb|ConvBlock|($c=512$)$\times2$\\
    %\hline
    &(512,1,1)   &Global Average Pooling\\
    \hline
    $34$&10         &Linear\\
    \hline
    \hline
    &$2.11\times10^7$&Number of Parameters\\
    \hline
  \end{tabular}
\end{table*}

\section{Discussions}

Table~\ref{tbl:compare-init} compares the proposed initialization method with the two existing methods, the Kaiming and Xavier methods. 
The proposed method is an extension of the Kaiming method by including all basic components in the model, as shown below. First we discuss the forward method. Here we assume that the convolution operation performs without padding. Let the kernel size be $k\times k\times d$. Then, $|\vs(\ell,i)|=k^{2}d$. Because the Kaiming method ignores the pooling layer, we further assume that the structure does not have any pooling layer.  This can be realized in our model by setting all $g_{\ell}$ to the identity function, yielding $T_{\ell}=1$ and thereby $\tau_{\ell}=1/2$. Under these assumptions, we can then re-write the forward initialization method~\eqref{eq:init-forward} as
\begin{tsaligned}
  \sigma^2_{w^{(\ell)}}
  =\frac{M^\prime_\ell}{\tau_{\ell-1}\varepsilon_\ell}
  =\frac{2M^\prime_\ell}{M^\prime_\ell\cdot k^2d}
  =\frac{2}{k^2d}
\end{tsaligned}
This concludes that the proposed forward method coincides with the Kaiming's forward method. 
For the backward method, we use a similar assumption to fit our model to the Kaiming's simplified model. Let $d'$ be the number of channels of the next layer. Then, $|\vj(\ell,i)|=k^2d^\prime$ and $\gamma_\ell=1/2$. We observe that the number of forward connections equals to the number of backward connections, we have $\varepsilon_\ell=\sum_{i=1}^{M_{\ell-1}}|\vj(\ell,i)|$. Under these assumptions, the backward method can be rewritten as
\begin{tsaligned}
  \sigma^2_{w^{(\ell)}}
  =\frac{M_{\ell-1}}{\gamma_\ell\varepsilon_\ell}
  =\frac{2M_{\ell-1}}{M_{\ell-1}k^2d^\prime}
  =\frac{2}{k^2d^\prime}
\end{tsaligned}
implying that our backward method is equal to the Kaiming's backward method.

The proposed methods behaves as follows. 
For a convolution layer with padding, a less number of signals are fed to the convolution operation applied to the corner or the border of images. More precisely, the cardinality of $\vs(\ell,i)$ for the corner or the border is smaller than that for other areas, and so $\varepsilon_{\ell}$, which increases $\sigma_{w^{(\ell)}}^{2}$. Meanwhile, the Kaiming method does not consider the padding operation, which causes the difference between the proposed and Kaiming methods.

We next observe the behaviors of the two types of pooling layers in the forward analysis. For a layer equipped with the max pooling, a larger $T_{\ell}$ yields a larger $\tau_{\ell}$, making smaller $\sigma_{w^{(\ell)}}^{2}$ of the proposed initialization method~\eqref{eq:init-forward}. Actually, the maximum of multiple random variables with zero mean has a larger second moment than that of the individual random variable. The proposed forward method can suppress the effect by setting $\sigma_{w^{(\ell)}}^{2}$ to a smaller value. 
For a layer having the average pooling, a larger $T_{\ell}$ gets $\tau_{\ell}$ smaller and thereby the variance parameter determined by the initialization method, say $\sigma_{w^{(\ell)}}^{2}$, is increased. In general, taking the average of random variables with zero mean yields a smaller signal variance. The proposed forward method compensates this effect by giving a larger variance parameter $\sigma_{w^{(\ell)}}^{2}$. 

We finally see how the proposed backward method behaves for the two types of pooling layers. For the max pooling operation, as $T_{\ell}$ is larger, $\tau_{\ell}$ is smaller, increasing $\sigma_{w^{(\ell)}}^{2}$ as in \eqref{eq:init-backward}. In the backward propagation, only the winner unit propagates the signal to the shallower layers, that decreases the signal variance in the shallower layer. To keep the backward signal variance to be equal to one, the proposed backward method gives a larger variance parameter $\sigma_{w^{(\ell)}}^{2}$. 
For the average pooling operation, the proposed backward method increases $\sigma_{w^{(\ell)}}^{2}$ as $T_{\ell}$ increases, because a larger $T_{\ell}$ increases a smaller $\gamma_{\ell}$. For the average pooling operation, a backward signal is divided evenly and sent back to the units in the previous layer, decreasing the backward signal variance. The proposed backward method compensates this effect by setting $\sigma_{w^{(\ell)}}^{2}$ to a larger value.

\begin{table*}[t!]
  % version: task.g2-backup 2020/08/03
  \centering
  \caption{Validation accuracies of 34-layer architecture using different initialization methods for various learning rates. For each learning rate $lr$, each initialization method provides an initial value for the model. The numerical values in the table represent the highest accuracy rate in a span of 1000 epochs. The numbers in italics indicate the best value in each initialization method and the bold face numbers represent the best performance for the entire table.
  	\label{tab:acc.g2}}
  %% \label{tbl:results}
  \begin{tabular}{l}
    (a) Car
    \\
    \begin{tabular}{|c|c|c|c|c|c|}
      \hline
      34-layer &
      \multicolumn{5}{|c|}{Initialization Methods} \\
      \hline
      Learning & \multirow{2}{*}{Xavier}  &Kaiming  &Kaiming   & ASV & ASV \\
      Rate ($lr$)&         &(forward)&(backward)&(forward)&(backward)\\
      \hline
      \hline
      $10^{-3}$&15.78&15.78&15.78&15.78&15.78\\
      $10^{-4}$&15.78&15.78&15.78&15.78&\textbf{\textit{81.49}}\\
      $10^{-5}$&\textit{71.95}&\textit{70.52}&\textit{73.10}&\textit{72.74}&63.85\\
      $10^{-6}$&40.17&52.58&50.93&51.15&52.30\\
      \hline
    \end{tabular}
    \\ \\
    (b) Food
    \\
    \begin{tabular}{|c|c|c|c|c|c|}
      \hline
      34-layer &
      \multicolumn{5}{|c|}{Initialization Methods} \\
      \hline
      Learning & \multirow{2}{*}{Xavier}   &Kaiming  &Kaiming   & ASV & ASV \\
      Rate ($lr$)& 						   &(forward)&(backward)&(forward)&(backward)\\
      \hline
      \hline
      $10^{-3}$&10.60&10.60&10.60&10.60&10.60\\
      $10^{-4}$&69.36&\textit{75.72}&69.36&\textit{76.49}&\textbf{\textit{78.81}}\\
      $10^{-5}$&\textit{72.83}&69.17&\textit{69.75}&69.75&66.28\\
      $10^{-6}$&60.89&65.51&66.28&65.13&62.81\\
      \hline
    \end{tabular}
    \\ \\
    (c) Fungi
    \\
    \begin{tabular}{|c|c|c|c|c|c|}
      \hline
      34-layer &
      \multicolumn{5}{|c|}{Initialization Methods} \\
      \hline
      Learning & \multirow{2}{*}{Xavier}  &Kaiming  &Kaiming   & ASV & ASV \\
      Rate ($lr$)&         &(forward)&(backward)&(forward)&(backward)\\
      \hline
      \hline
      $10^{-3}$&24.61&24.61&24.61&24.61&24.61\\
      $10^{-4}$&\textit{65.23}&\textit{68.16}&66.02&\textit{67.97}&\textbf{\textit{69.73}}\\
      $10^{-5}$&62.11&65.62&\textit{66.99}&64.06&64.45\\
      $10^{-6}$&52.93&59.77&56.45&61.33&56.84\\
      \hline
    \end{tabular}
    \\
  \end{tabular}
\end{table*}
\begin{figure}
  \centering
  \begin{tabular}{l}
    \includegraphics[width=0.9\linewidth]{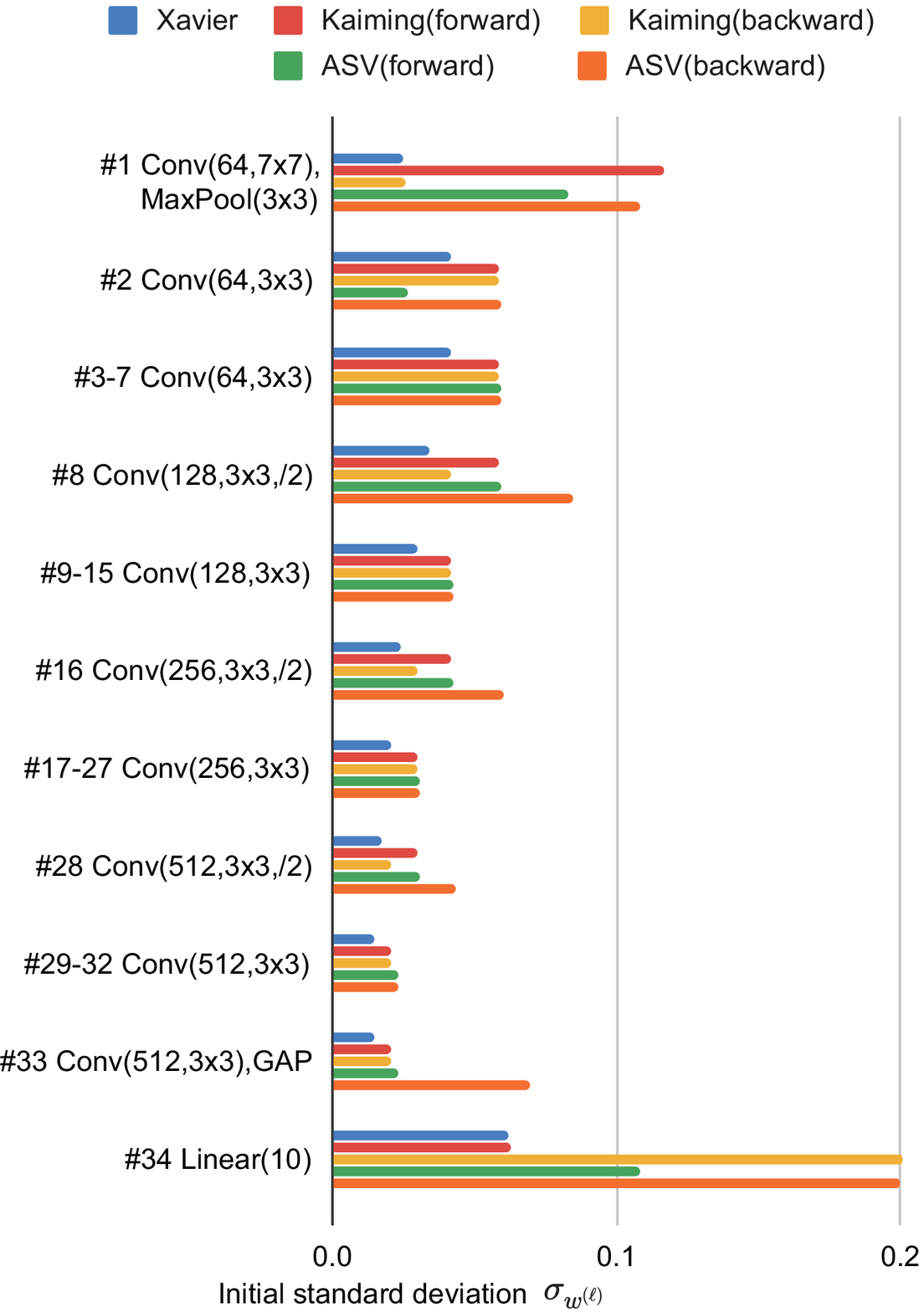}
  \end{tabular}
  \caption{Graph comparing parameters $\sigma_{w^{(\ell)}}$ used for initialization for 34-layer architecture. The number after the $\#$ represents the layer number $\ell$, with some layers grouped together. It can be seen that the proposed backward method gives a larger variance, especially on the layers where a stride or pooling is applied wherein in these layers, the resolution of the feature map changes. \label{fig:initstd}}
\end{figure}
\section{Experiments}
We carried out computational experiments to investigate how the differences of the initialization methods affect the stochastic gradient descent of deep convolutional structures. 

The proposed initialization methods were examined on three public datasets: Car, Food, and Fungi. Each of the three datasets are divided into ten categories yielding ten output units in the last layer of the deep convolutional layer for multi-category classification. The three datasets comprise three-channel color images. The images were clipped, resized to $224\times 224$ pixels and normalized to allow us to examine the proposed method using exactly same neural architectures (Table~\ref{tab:arch}) for the three different datasets. Data augmentation was not performed in the gradient descent process. The first dataset Car is available in VMMRdb~\cite{vmmrdb2017}. Seventy-five percent of the data was randomly selected for training and the rest was used for validation. The second dataset Food is a subset of iFood 2019~\cite{kaur2019foodx}. The dataset was divided into two datasets of 90\% and 10\%, each of which was used for training and validation, respectively. The last dataset Fungi was prepared by extracting the super-category Fungi in the hierarchical categorization of iNaturalist 2019 in ~\cite{iNaturalistdb} and its exclusive subsets of 75\% and 25\% were used for training and validation. 

As shown in Table~\ref{tab:arch}, the model architecture used in the experiments on the above three datasets has a 34-layer deep structure that consists of the max-pooling layers, the convolutional layer with padding, the global averaging layers, and the fully connected layer. The cross-entropy loss was used for the loss function. One thousand epochs of the gradient descent based on Adam were implemented with the mini-batch size 64. The learning rate was varied with four values $10^{-6},10^{-5},10^{-4},10^{-3}$. The two proposed methods were compared with the \textbf{Kaiming forward} method, \textbf{Kaiming backward} method, and \textbf{Xavier} method. We refer to the proposed forward and backward methods as \textbf{ASV forward} and \textbf{ASV backward} methods, respectively. For ASV backward method, learning is stabilized by taking the smaller value of the variance parameter \eqref{eq:init-backward} and three times the value derived without the pooling operations. The classification accuracies on the validation set were assessed during training, and the highest one is reported. ReLU function is used as the activation function in all layers and no activation function is used for the output layer.

Table~\ref{tab:acc.g2}a shows the classification accuracies on the dataset Car. ASV backward achieved the highest accuracy 81.49\%. The accuracy was recorded when using $10^{-4}$ as the learning rate. The accuracy obtained by each of the other methods with the learning rate of $10^{-5}$ was the highest among the accuracies with different learning rates. The proposed backward method successfully trained the deep network stably with a larger learning rate $10^{-4}$, although the other methods suffered from the plateau phenomenon at the same learning rate. Comparing the two ASV methods, the backward method performed better than the forward method on the dataset Car as well as on the other two datasets as shown in Table \ref{tab:acc.g2}b and \ref{tab:acc.g2}c. The backward method is derived so that the variances of the backward signals are maintained, which is a more direct approach for preventing the signal vanishing and the signal explosion compared to the forward method. We conjecture that the reason why the same phenomenon was not observed for the Kaiming forward and backward methods is because the two methods were derived from a too simplified variance model. A prominent difference between the ASV backward method and the ASV forward method appears when the resolution of the feature map is changed by the sub-sampling operation. Our theory suggests that a larger variance affects such layers better, and ASV backward method can make the variance larger (See Figure~\ref{fig:initstd}), which leads to better performance. Meanwhile, the Kaiming variance model ignores the sub-sampling operation. Kaiming backward method derived from the Kaiming variance model that cannot capture the variance change caused by the resolution change.

In addition to the 34-layer architecture, the initialization methods were tested with a deeper 50-layer architecture presented in Table~\ref{tab:arch-f50}. Details on the implementation of this experiment can be seen in Appendix~\ref{sec:addexp}. Table~\ref{tab:acc-f50} reports the validation accuracies on the dataset Car. ASV backward method performed best, although the accuracy is slightly worse than that from the 34-layer architecture. This suggests that the proposed method works successfully with deeper convolutional structures. To get a better performance with deeper structures, it may be required to combine the proposed methods with other heuristics to avoid overfitting such as data augmentation, although we did not implement them because the purpose of the experiments is investigation of the effects coming from different initialization methods by removal of other factors. 
\section{Conclusion and Future Work}
Using CNN in various image processing tasks requires practitioners to understand that proper initialization of the model parameters is crucial to the stability of learning. With a proper initialization method, it can determine the outcome of training and testing process. In this paper, we have shown a re-formulation of the CNN structure to introduce our proposed initialization methods which take into account all the components of CNN, including the ReLU activation function, fully connected layer, convolution layer and pooling operations. The proposed initialization methods, not only supports the CNN structure theoretically, these methods as well made significant improvements to the recognition performance demonstrated in the experimental results.

While ASV initialization method produces an impressive result under the CNN structure for these specific datasets, there are more neural network structures that the ASV initialization method has not supported yet theoretically. If the proposed initialization method were to be extended to other neural networks, it might show promising results.

{\small
\bibliographystyle{ieee_fullname}

}

\appendix
\onecolumn
\numberwithin{equation}{section}
\section{Tensor Representation of CNN}
\label{sec:detailed-formulation}
CNN structures for image analysis are usually represented with three-mode tensors. In this section, it is seen that our CNN model is a vectorization of the standard structures of CNN.
\begin{definition-waku}[Vectorization]
  \label{def:formulation-forward}
  Let $\vT\in\bR^{n_1\times\dots\times n_d}$. 
  With a bijection map $\mu:[n_1]\times\dots\times[n_d]\to[n_1\dots n_d]$, 
  a vector $\vt := \left[t_{1},\dots,t_{n_{1}\dots n_{d}}\right]^\top\in\bR^{n_1\dots n_d}$ defined as
  \begin{tsaligned}
    t_{i}=T_{\mu^{-1}(i)} 
  \end{tsaligned}
  is said to be a vectorization of $\vT$, and denoted by
  \begin{tsaligned}
    \vt=\vect\vT. 
  \end{tsaligned}
  A bijection map $\mu$ is usually defined so that
  \begin{tsaligned}\label{eq:stdmu-in-vect-def}
    \vect(\vT)=\begin{bmatrix}
      T_{1,1,\ldots ,1}&T_{2,1,\ldots ,1}&\cdots &T_{n_{1},1,\ldots ,1}
      &T_{1,2,1,\ldots ,1}&\cdots &T_{n_1,n_2,\ldots,n_d}
    \end{bmatrix}^\top. 
  \end{tsaligned}
\end{definition-waku}
In case of $d=2$, the expression \eqref{eq:stdmu-in-vect-def}
conforms to a standard vectorization of matrices.

Unlike the fully connected layer that receives all signals in the previous layer, the inputs of the convolution and the pooling operations are a part of signals that are expressed with a sub-vector of the vectorization of a feature map. In Section~\ref{sec:cnnmdl}, we have introduced a notation to denote a sub-vector. 
\begin{definition-waku}[Sub-Vector]
  Let $\va\in\bR^{n}$. Let $\vs=\{s_1,\dots,s_m\}\subseteq[n]$
  be an ordered set such that
  $\forall i,j\in[m],i<j \implies s_i<s_j)$.
  Notation $\va_{\vs}$ expresses a sub-vector of $\va$ such that
  \begin{equation*}
    \va_{\vs}=\begin{bmatrix}a_{s_1}&\dots&a_{s_m}\end{bmatrix}^\top\in\bR^m. 
  \end{equation*}
\end{definition-waku}

In what follows, we describe how our CNN model is derived from the standard CNN structures.

\subsection{Convolution}\label{subsec:formulation:conv}
Consider a two-dimensional convolution operation with zero padding. The convolution may have a stride. We denote $(k_{w}, k_{h})$, $(p_{w}, p_{h})$, and $(s_{w}, s_{h})$ as the kernel size, the padding size, and the stride size respectively. Suppose that $\Z^{(\ell-1)}\in\bR^{w\times h\times d}$ is given as an input of convolution. The input in which the zero padding is applied is given by $\vP^{(\ell)}\in\bR^{(w+2p_w)\times(h+2p_h)\times d}$ with
\begin{tsaligned}
  P^{(\ell)}_{i,j,k}:=
  \begin{cases}
    Z^{(\ell-1)}_{i-p_w,j-p_h,k}&\text{if}~1\leq i-p_w\leq w~\text{and}~1\leq j-p_h\leq h\\
    0&\text{otherwise}. 
 \end{cases}
\end{tsaligned}
Denote by $\vW^{(\ell)}_i\in\bR^{k_w\times k_h\times d}~(i=1,\dots,d^\prime)$ and
$\vb^{(\ell)}\in\bR^{d^\prime}$ the two-dimensional convolution kernel and the bias parameter, where $d$ and $d^{\prime}$, respectively, are the numbers of channels of the previous layer and the current layer. The feature map transformed with the convolution kernel is expressed as
$\vU^{(\ell)}\in\bR^{w^\prime\times h^\prime\times d^\prime}$
with
\begin{tsaligned}
  U^{(\ell)}_{i,j,k}:=b^{(\ell)}_k+
    \sum_{\xi_1=1}^{k_w}\sum_{\xi_2=1}^{k_h}\sum_{\xi_3=1}^d
      W^{(\ell)}_{k,\xi_1,\xi_2,\xi_3}
      P^{(\ell)}_{s_w(i-1)+\xi_1,s_h(j-1)+\xi_2,\xi_3}.
  \label{eq:def:conv2d-naive}
\end{tsaligned}
where $(w^{\prime},d^{\prime})$ is the size of the output of the convolution:
\begin{tsaligned}
  w^\prime&=\left\lfloor\frac{w+2p_w-k_w}{s_w}\right\rfloor+1,&
  h^\prime&=\left\lfloor\frac{h+2p_h-k_h}{s_h}\right\rfloor+1.
\end{tsaligned}
Letting
\begin{tsaligned}
  \lambda_{w,i}&:=s_w(i-1)+\xi_1-p_w,&
  \lambda_{h,j}&:=s_h(j-1)+\xi_2-p_h, 
\end{tsaligned}
from the definition of $\vP$, we have
\begin{tsaligned}
  1\leq\lambda_{w,i}\leq w\,
  \mbox{ and }
  1\leq\lambda_{h,j}\leq h
  \implies
  P^{(\ell)}_{\lambda_{w,i}+p_w,\lambda_{h,j}+p_h,k}
  =Z^{(\ell-1)}_{\lambda_{w,i},\lambda_{h,j},k}
\end{tsaligned}
Define index sets as 
\begin{tsaligned}
  \cA(\ell;i,j)&:=\{(\xi_1,\xi_2,\xi_3)\in[k_w]\times[k_h]\times[d]
    \mid1\leq\lambda_{w,i}\leq w,1\leq\lambda_{h,i}\leq h\},\\
  \cS(\ell;i,j)&:=\{(\lambda_{w,i},\lambda_{h,j},\xi_3)\mid(\xi_1,\xi_2,\xi_3)\in\cA(\ell;i,j)\}. 
\end{tsaligned}
Notice that the two sets have a one-to-one correspondence.
The set of the corresponding pairs are denoted by
$(\cA,\cS)\in(\cA(\ell;i,j),\cS(\ell;i,j))$. 
Then, the two-dimensional convolution
\eqref{eq:def:conv2d-naive} can be rewritten as
\begin{tsaligned}
U^{(\ell)}_{i,j,k}=b^{(\ell)}_k+\sum_{(\cA,\cS)\in(\cA(\ell;i,j),\cS(\ell;i,j))}
W^{(\ell)}_{k,\cA}Z^{(\ell-1)}_{\cS}. 
\end{tsaligned}
Transforming the index sets as 
\begin{tsaligned}
  \va(\ell,t)&:=\{\mu(\cA)\mid(i,j,k)=\mu^{-1}(t),\cA\in\cA(\ell;i,j)\},
  \\
  \vs(\ell,t)&:=\{\mu(\cS)\mid(i,j,k)=\mu^{-1}(t),\cS\in\cS(\ell;i,j)\},
  \\
  c(\ell,t)&:=k~\text{s.t.}~(i,j,k)=\mu^{-1}(t)
\end{tsaligned}
and vectorizing the tensors
\begin{tsaligned}
  \z^{(\ell-1)}&=\vect\Z^{(\ell-1)},&
  \vu^{(\ell)}&=\vect\vU^{(\ell)},&
  \vw^{(\ell)}_i&=\vect\vW^{(\ell)}_i, 
\end{tsaligned}
we have an alternative expresson based on vectorizations:  
\begin{tsaligned}
  u^{(\ell)}_i
  &=b^{(\ell)}_{c(\ell,i)}+\sum_{(a,s)\in(\va(\ell,i),\vs(\ell,i))}
    w^{(\ell)}_{c(\ell,i),a}z^{(\ell-1)}_s\\
  &=b^{(\ell)}_{c(\ell,i)}+\ip{\vw^{(\ell)}_{c(\ell,i),\va(\ell,i)},
    \z^{(\ell-1)}_{\vs(\ell,i)}}. 
\end{tsaligned}
We also have 
\begin{tsaligned}
  M_{\ell-1}&=w\cdot h\cdot d,&
  M^\prime_\ell&=w^\prime\cdot h^\prime\cdot d^\prime,&
  C_\ell&=d^\prime,&
  S_\ell&=k_w\cdot k_h\cdot d. 
\end{tsaligned}

For the back propagation, we prepare another
backward convolution kernel
$\vWtil^{(\ell)}_i\in\bR^{k_w\times k_h\times d^\prime}~(i=1,\dots,d)$
from the forward convolution kernel
$\vW^{(\ell)}_i\in\bR^{k_w\times k_h\times d}~(i=1,\dots,d^\prime)$
by re-shaping the tensor and re-ordering the entries as
\begin{tsaligned}
  \tilde{W}^{(\ell)}_{i,j,k,l}=W^{(\ell)}_{l,j,k,i}
\end{tsaligned}
Then, it holds that 
\begin{tsaligned}
  \pderiv{U^{(\ell)}_{i,j,k}}{Z^{(\ell-1)}_{l,m,n}}\neq0
  &\implies(l,m,n)\in\cS(\ell;i,j)\\
  &\Longleftrightarrow\exists!(\zeta_1,\zeta_2,n)\in\cA(\ell;i,j),~
    \pderiv{U^{(\ell)}_{i,j,k}}{Z^{(\ell-1)}_{l,m,n}}
    =W^{(\ell)}_{k,\zeta_1,\zeta_2,n}
    =\tilde{W}^{(\ell)}_{n,\zeta_1,\zeta_2,k}
\end{tsaligned}
where
$\zeta_{1}$ and $\zeta_{2}$ are defined as 
\begin{tsaligned}
  \zeta_1&=l+p_w-s_w(i-1),&
  \zeta_2&=m+p_h-s_h(j-1)
\end{tsaligned}
Letting two index sets as
\begin{tsaligned}
  \cJ(\ell;l,m)&:=\{(i,j,k)\in[w^\prime]\times[h^\prime]\times[d^\prime]
    \mid n\in[d],(l,m,n)\in\cS(\ell;i,j)\}\\
    &=\{(i,j,k)\in[w^\prime]\times[h^\prime]\times[d^\prime]
    \mid1\leq\zeta_1\leq k_w,1\leq\zeta_2\leq k_h\},
    \\
  \cH(\ell;l,m)&:=\{(\zeta_1,\zeta_2,k)
    \mid(i,j,k)\in\cJ(\ell;l,m)\}, 
\end{tsaligned}
we have
\begin{tsaligned}
  \pderiv{E}{Z^{(\ell-1)}_{l,m,n}}
  &=\sum_{(i,j,k)\in[w^\prime]\times[h^\prime]\times[d^\prime]}
    \pderiv{E}{U^{(\ell)}_{i,j,k}}\pderiv{U^{(\ell)}_{i,j,k}}{Z^{(\ell-1)}_{l,m,n}}
  =\sum_{(\cJ,\cH)\in(\cJ(\ell;l,m),\cH(\ell;l,m))}
    \pderiv{E}{U^{(\ell)}_\cJ}\tilde{W}^{(\ell)}_{n,\cH}. 
\end{tsaligned}
Furthermore, letting
\begin{tsaligned}
  &J_\ell:=k_w\cdot k_h\cdot d^\prime,
  \qquad
  \tilde{C}_\ell:=d,
  \\
  &\vj(\ell,t):=\{\mu(\cJ)\mid(l,m,n)=\mu^{-1}(t),\cJ\in\cJ(\ell;l,m)\},
  \\
  &\vh(\ell,t):=\{\mu(\cH)\mid(l,m,n)=\mu^{-1}(t),\cH\in\cH(\ell;l,m)\},
  \\
  &\tilde{c}(\ell,t):=n~\text{s.t.}~(l,m,n)=\mu^{-1}(t),
  \\
  &\vwtil^{(\ell)}_i:=\vect\vWtil^{(\ell)}_i,
  \qquad
  \Delta\z^{(\ell-1)}:=\vect\pderiv{E}{\Z^{(\ell-1)}},
  \qquad
  \Delta\vu^{(\ell)}:=\vect\pderiv{E}{\vU^{(\ell)}},
\end{tsaligned}
we obtain
\begin{tsaligned}
  \Delta z^{(\ell-1)}_i
  =\sum_{(j,h)\in(\vj(\ell,i),\vh(\ell,i))}
    \Delta u^{(\ell)}_{j}
    \tilde{w}^{(\ell)}_{\tilde{c}(\ell,i),h}
  =\ip{\Delta\vu^{(\ell)}_{\vj(\ell,i)},
    \vwtil^{(\ell)}_{\tilde{c}(\ell,i),\vh(\ell,i)}}. 
\end{tsaligned}

\subsection{Fully Connected Layer}
The affine transformation used in the fully connected layer is a special case of the convolution operation. Let $M_{\ell-1}$ and $M_{\ell}^{\prime}$ be the number of input units and output units for the fully connected layer.
When a feature map $\z^{(\ell-1)}\in\bR^{M_{\ell-1}}$ are received,
the affine transformation with weight parameters  
$\vw^{(\ell)}_i\in\bR^{M_{\ell-1}}~(i=1,\dots,M^\prime_\ell)$
and bias parameters $\vb^{(\ell)}\in\bR^{M^\prime_\ell}$
produces
\begin{tsaligned}
  u^{(\ell)}_{i}=\ip{\vw^{(\ell)}_i,\z^{(\ell-1)}}+b^{(\ell)}_i
\end{tsaligned}
which is in the class of the forward propagation
\eqref{eq:def:layer-conv}
by setting
\begin{tsaligned}
  \va(\ell,i)&=[M_{\ell-1}],&
  \vs(\ell,i)&=[M_{\ell-1}],&
  c(\ell,i)&=i. 
\end{tsaligned}
Here, constants determining the size of vectors are given by 
$C_\ell:=M^\prime_\ell$ and 
$S_\ell:=M_{\ell-1}$. 

For back propagation,
the backward weight matrix
$\tilde{\vW}^{(\ell)}:=\left[\tilde{\vw}_{1}^{(\ell)},\dots,\tilde{\vw}_{M_{\ell-1}}^{(\ell)}\right]^\top\in\bR^{M^\prime_\ell\times M_{\ell-1}}$
is defined by transposing the forward weight
matrix
$\vW^{(\ell)}:=\left[\vw_{1}^{(\ell)},\dots,\vw_{C_{\ell}}^{(\ell)}\right]^\top\in\bR^{C_{\ell}\times S_{\ell}}$
as
\begin{tsaligned}
  \tilde{w}^{(\ell)}_{i,j}=w^{(\ell)}_{j,i}.  
\end{tsaligned}
Letting 
\begin{tsaligned}
  J_\ell&=M^\prime_\ell,&
  \tilde{C}_\ell&=M_{\ell-1},&
  \vj(\ell,i)&=[M^\prime_\ell],&
  \vh(\ell,i)&=[M^\prime_\ell],&
  \tilde{c}(\ell,i)&=i, 
\end{tsaligned}
Equation \eqref{eq:formulation:conv-backward} coincides with
the backward propagation operation in the fully connected layer.

\subsection{Activation function}
Activation functions are usually applied 
just after the convolution operation in the convolution layer or
the affine transformation in the fully connected layer.
Following the convolution operation or the affine transformation, 
an activation function is applied entry-wise
for $\vu^{(\ell)}\in\bR^{M^\prime_\ell}$ as 
\begin{tsaligned}
  v^{(\ell)}_i=f_\ell\left(u^{(\ell)}_i\right)
\end{tsaligned}
to produce $\vv^{(\ell)}\in\bR^{M^\prime_\ell}$.
Currently, the most popular activation function is ReLU
expressed as
$f_\ell(x)=\max\{0,x\}$. 
This formulation can also express layers having no activation by setting $f_{\ell}$ to the identity function.

For back propagation, if the activation function is differentiable,
\begin{tsaligned}
  \Delta u^{(\ell)}_i=\Delta v^{(\ell)}_i\cdot
    f^\prime_\ell\left(u^{(\ell)}_i\right)
\end{tsaligned}
where $f^\prime_\ell$ is the derivative of the activation function. 
Subgradients are used when the activation function is not differentiable.
\subsection{Pooling}
Pooling operations such as the max pooling and the average pooling aggregate each of divisions of a feature map to obtain a subsampled feature map. Suppose that input feature maps fed to the pooling operation are divided evenly and exclusively. Denote by $(t_{w},t_{h})\in\bN^{2}$ the pooling size. Given a feature map $\vV^{(\ell)}\in\bR^{w\times h\times d}$ just after activation, the max pooling operation transforms it as
\begin{tsaligned}\label{eq:def:pool2d-naive-maxpool}
  Z^{(\ell)}_{i,j,k}:=  
  \max_{\substack{1\leq\xi_1\leq t_w,\\1\leq\xi_2\leq t_h}}
  V^{(\ell)}_{t_w(i-1)+\xi_1,t_h(j-1)+\xi_2,k}
\end{tsaligned}
and the average operation transforms the feature map $\vV^{(\ell)}$ as
\begin{tsaligned}\label{eq:def:pool2d-naive-avepool}
  Z^{(\ell)}_{i,j,k}:=
    \frac{1}{t_wt_h}\sum_{\xi_1=1}^{t_w}\sum_{\xi_2=1}^{t_h}
      V^{(\ell)}_{t_w(i-1)+\xi_1,t_h(j-1)+\xi_2,k}. 
\end{tsaligned}
The size of the transformed feature map, $(w^{\prime},h^{\prime})$, is given by
\begin{tsaligned}
  w^\prime&:=\left\lfloor\frac{w-t_w}{t_w}\right\rfloor+1,&
  h^\prime&:=\left\lfloor\frac{h-t_h}{t_h}\right\rfloor+1.
\end{tsaligned}
Let $T_\ell:=t_w\cdot t_h$, and define an index set
\begin{tsaligned}
  \cT(\ell;i,j,k):=\{(t_w(i-1)+\xi_1,t_h(j-1)+\xi_2,k)\mid\xi_1\in[t_w],\xi_2\in[t_h]\}, 
\end{tsaligned}
the two types of pooling operations described
in \eqref{eq:def:pool2d-naive-maxpool} and \eqref{eq:def:pool2d-naive-avepool},
respectively, 
are rearranged as
\begin{tsaligned}
 Z^{(\ell)}_{i,j,k}=
 \max_{\cT\in\cT(\ell;i,j,k)}V^{(\ell)}_\cT
\end{tsaligned}
and
\begin{tsaligned}
 Z^{(\ell)}_{i,j,k}=
   \frac{1}{T_\ell}\sum_{\cT\in\cT(\ell;i,j,k)}V^{(\ell)}_\cT. 
\end{tsaligned}
Letting
\begin{tsaligned}
  \z^{(\ell)}&=\vect\Z^{(\ell)},&
  \vv^{(\ell)}&=\vect\vV^{(\ell)},&
  \vt(\ell,t)&=\{\mu(\cT)\mid(i,j,k)=\mu^{-1}(t),\cT\in\cT(\ell;i,j,k)\}
\end{tsaligned}
and defining the pooling function
$g_\ell:\bR^{T_\ell}\to\bR$
for the max pooling and
the average pooling, respectively, as 
\begin{tsaligned}
  g_\ell(\x):=
  \max_{i=1,\dots,T_\ell}x_i
\end{tsaligned}
and
\begin{tsaligned}
 g_\ell(\x):=
 \frac{1}{T_\ell}\sum_{i=1}^{T_\ell}x_i, 
\end{tsaligned}
we obtain a vectorized pooling expression as
\begin{tsaligned}
  z^{(\ell)}_i=g_\ell\left(\vv^{(\ell)}_{\vt(\ell,i)}\right). 
\end{tsaligned}

Finally, we discuss the pooling operations in the back propagation. For any $(i,j,k)\in[w]\times[h]\times[d]$, there exists $(l,m)\in[w^\prime]\times[h^\prime]$ uniquely such that $(i,j,k)\in\cT(\ell;l,m)$. That is because $\cT(\ell;i,j)$ divides the received feature map exclusively.  We define a mapping function $\cD(\ell;i,j,k)$ as
\begin{tsaligned}
  \cD(\ell;i,j,k):=\left(\left\lfloor\frac{i-1}{t_w}\right\rfloor+1,
    \left\lfloor\frac{j-1}{t_h}\right\rfloor+1,k\right)
\end{tsaligned}
that provides another representation of the relationship of
$(i,j,k)\in\cT(\ell;l,m)$.
Letting
\begin{tsaligned}
  d(\ell,t)&:=\mu(D(\ell;i,j,k))~\text{where}~(i,j,k)=\mu^{-1}(t),\\
  \Delta\z^{(\ell)}&:=\vect\pderiv{E}{\Z^{(\ell)}}, \qquad
  \Delta\vv^{(\ell)}:=\vect\pderiv{E}{\vV^{(\ell)}},\\
\end{tsaligned}
we have a vectorized expression of tensors as 
\begin{tsaligned}
  \Delta v^{(\ell)}_i&=\Delta z^{(\ell)}_{d(\ell,i)}
    \pderiv{z^{(\ell)}_{d(\ell,i)}}{v^{(\ell)}_i}. 
\end{tsaligned}
The gradient
$\pderiv{z^{(\ell)}_{d(\ell,i)}}{v^{(\ell)}_i}$ in the above
expression is obtained from the fact 
\begin{tsaligned}
  \pderiv{g_\ell(\x)}{x_i}=\begin{cases}
    \delta_{x_i,g_\ell(\x)}&\text{if $g_{\ell}$ performs the max pooling},\\
    \frac{1}{T_\ell}&\text{if $g_{\ell}$ performs the average pooling}.\\
  \end{cases}
\end{tsaligned}

\section{Variance Propagation}
\label{sec:variance}
In this section, we derive the recursive expressions of forward variances and backward variances, \eqref{eq:var-forward} and \eqref{eq:var-backward}, from our CNN model defined in \eqref{eq:def:layer-conv}, \eqref{eq:def:layer-act}, and \eqref{eq:def:layer-pool}. We start the analysis from the assumption that the model parameters are drawn from $w^{(\ell)}_{i,j}\sim\cN(0,\sigma^2_{w^{(\ell)}})$ and $b^{(\ell)}_i\sim\cN(0,\sigma^2_{b^{(\ell)}})$ as described in \eqref{eq:init-probmdl}. Then, their first and second moments are immediately obtained as
\begin{tsaligned}
  \bE_{x}\left[w^{(\ell)}_{i,j}\right]&=0,\qquad
  \bE_{x}\left[b^{(\ell)}_{i}\right]=0,\\
  \bE_{x}\left[\left(w^{(\ell)}_{i,j}\right)^2\right]
  &=
  \bE_{x}\left[\left(w^{(\ell)}_{i,j}\right)^2\right]
  -
  \left(\bE_{x}\left[w^{(\ell)}_{i,j}\right]\right)^2
  =
  \text{Var}[w^{(\ell)}_{i,j}]=\sigma^2_{w^{(\ell)}},
  \\
  \bE_{x}\left[\left(b^{(\ell)}_i\right)^2\right]
  &=
  \bE_{x}\left[
    \left(b^{(\ell)}_i\right)^2
    \right]
  -\left(\bE_{x}\left[b^{(\ell)}_i\right]\right)^2
  =\text{Var}[b^{(\ell)}_i]=\sigma^2_{b^{(\ell)}}.
\end{tsaligned}
Our analysis shall use the following lemma:
\begin{lemma-waku}\label{lem:dist-max}
Let $x_{1},\dots,x_{n}$ be random variables independently and identically distributed according to a density function~$p$. Then, $\hat{x}=\max_{i\in[n] }x_{i}$ is drawn according to a density function $\hat{p}$ defined as 
  \begin{tsaligned}
    \hat{p}(x):=np(x)\left(\int_{-\infty}^xp(s)d s\right)^{n-1}.
  \end{tsaligned}
\end{lemma-waku}

\paragraph*{Proof for Lemma~\ref{lem:dist-max}: }
  Let $F$ and $\hat{F}$ be the cumulative density functions of $p$ and $\hat{p}$, respectively. Then, we have 
  \begin{tsaligned}
    \hat{F}(x)
    =\Pr(\hat{x}\leq x)
    =\Pr(x_1\leq x, \dots , x_n\leq x)
    % ($x_{i}$ are independent)
    =\prod_{i=1}^n\Pr(x_i\leq x)
    =F(x)^n
  \end{tsaligned}
  yielding 
\begin{tsaligned}
    \hat{p}(x)
    =\frac{d\hat{F}}{d x}
    =nF(x)^{n-1}\frac{d F}{d x}
    =n\left(\int_{-\infty}^xp(s)d s\right)^{n-1}p(x).
\end{tsaligned}
\qed

In the following two subsections~\ref{sec:appx-forward-variance} and
~\ref{sec:appx-backward-variance}, 
the two recursive expressions for variances propagations, \eqref{eq:var-forward} and \eqref{eq:var-backward}, are derived,
respectively. 
\subsection{Derivation of forward variance propagation~\eqref{eq:var-forward}}
\label{sec:appx-forward-variance}
Assume that each entry in $\vu^{(\ell-1)}$ is iid according to
$\cN(0,q^{(\ell-1)})$ for $\ell=1,\dots,L-1$.
Consider the forward signal variance in $\ell$th layer,
$q^{(\ell)}_{i}=\Var{u^{(\ell)}_{i}}$. 
Since $w^{(\ell)}_{i,j}$ and $b^{(\ell)}_i$ are drawn independently, 
the expectation of the forward signal~$u^{(\ell)}_{i}$
can be expressed as 
\begin{align*}
  \Ex{u^{(\ell)}_i}
  &=\Ex{\ip{\vw^{(\ell)}_{c(\ell,i),\va(\ell,i)},\z^{(\ell-1)}_{\vs(\ell,i)}}
    +b^{(\ell)}_{c(\ell,i)}}\\
  &=\left(\sum_{(a,s)\in(\va(\ell,i),\vs(\ell,i))}
    \Ex{w^{(\ell)}_{c(\ell,i),a}}\Ex{z^{(\ell-1)}_s}\right)
    +\Ex{b^{(\ell)}_{c(\ell,i)}}
  =0. 
\end{align*}
We observe that 
\begin{tsaligned}
  \Ex{\left<\vw^{(\ell)}_{c(\ell,i),\va(\ell,i)},\z^{(\ell-1)}_{\vs(\ell,i)}\right>^2}
  &=\sum_{(a,s)\in(\va(\ell,i),\vs(\ell,i))}
    \Ex{\left(w^{(\ell)}_{c(\ell,i),a}\right)^2}
    \Ex{\left(z^{(\ell-1)}_s\right)^2}\\
  &\qquad+\sum_{\substack{(a,s),(a^\prime,s^\prime)\in(\va(\ell,i),\vs(\ell,i)),\\
    a\neq a^\prime,s\neq s^\prime}}
    \Ex{w^{(\ell)}_{c(\ell,i),a}}
    \Ex{w^{(\ell)}_{c(\ell,i),a^\prime}}
    \Ex{z^{(\ell-1)}_sz^{(\ell-1)}_{s^\prime}}\\
    &=\sigma^2_{w^{(\ell)}}\sum_{s\in\vs(\ell,i)}\Ex{\left(z^{(\ell-1)}_s\right)^2},
\end{tsaligned}
and that
\begin{tsaligned}
  &\Ex{\ip{\vw^{(\ell)}_{c(\ell,i),\va(\ell,i)},\z^{(\ell-1)}_{\vs(\ell,i)}}b^{(\ell)}_{c(\ell,i)}}
  =\Ex{\ip{\vw^{(\ell)}_{c(\ell,i),\va(\ell,i)},\z^{(\ell-1)}_{\vs(\ell,i)}}}
    \Ex{b^{(\ell)}_{c(\ell,i)}}
  =0
\end{tsaligned}
to express the signal variance $q^{(\ell)}_i$ as 
\begin{tsaligned}
  q^{(\ell)}_i
  &=\Ex{\left(u^{(\ell)}_i\right)^2}-\Ex{u^{(\ell)}_i}^2
  =\Ex{\left(\ip{\vw^{(\ell)}_{c(\ell,i),\va(\ell,i)},\z^{(\ell-1)}_{\vs(\ell,i)}}
    +b^{(\ell)}_{c(\ell,i)}\right)^2}
  \\
  &=\Ex{\ip{\vw^{(\ell)}_{c(\ell,i),\va(\ell,i)},\z^{(\ell-1)}_{\vs(\ell,i)}}^2}
  +
  2\Ex{\ip{\vw^{(\ell)}_{c(\ell,i),\va(\ell,i)},\z^{(\ell-1)}_{\vs(\ell,i)}}b^{(\ell)}_{c(\ell,i)}}
    +\Ex{\left(b^{(\ell)}_{c(\ell,i)}\right)^2}\\
  &=\sigma^2_{b^{(\ell)}}+\sigma^2_{w^{(\ell)}}
    \sum_{s\in\vs(\ell,i)}\Ex{\left(z^{(\ell-1)}_s\right)^2}. 
  \label{eq:variance-forward-base}
\end{tsaligned}

The statistics $\Ex{\left(z^{(\ell-1)}_s\right)^2}$ depends on the activation function and the pooling operation in $(\ell-1)$th layer. We here assume that the activation function is ReLU.
Equation~\eqref{eq:variance-forward-base} implies
that, to show the equation \eqref{eq:var-forward},
it suffices to derive
\begin{tsaligned}
  \sum_{s\in\vs(\ell,i)}\Ex{\left(z^{(\ell-1)}_s\right)^2}
  =
  q^{(\ell-1)}\tau_{\ell-1}S_{\ell,i}
\end{tsaligned}
for the two cases in which the max pooling and the average pooling, respectively, are used for the pooling operation. 

\subsubsection{Case for the max pooling}
Let
\begin{tsaligned}
  \label{eq:maxpool-maximum-element}
  \hat{u}^{(\ell-1)}_s:=\max_{t\in\vt(\ell-1,s)}u^{(\ell-1)}_{t}. 
\end{tsaligned}
The signal can be expressed as 
\begin{tsaligned}
  z^{(\ell-1)}_s=\max_{t\in\vt(\ell-1,s)}\max\left\{0,u^{(\ell-1)}_t\right\}
  =\max\left\{0,\max_{t\in\vt(\ell-1,s)}u^{(\ell-1)}_t\right\}
  =\max\left\{0,\hat{u}^{(\ell-1)}_s\right\}.
\end{tsaligned}
Let $\beta^{(\ell-1)}:=1/\sqrt{q^{(\ell-1)}}$. 
From Lemma~\ref{lem:dist-max}, 
the densities of the random variable $\hat{u}^{(\ell-1)}_s$ are
written as
\begin{tsaligned}
  \hat{p}_{\ell-1}(x)
  &=
  T_{\ell-1}\cN(x;0,q^{(\ell-1)})
  \left(\int_{-\infty}^x\cN(s;0,q^{(\ell-1)})\diff s\right)^{T_{\ell-1}-1}
  \\
  &=
  T_{\ell-1}\beta^{(\ell-1)}\phi(\beta^{(\ell-1)}x)
  \left(\int_{-\infty}^x
  \beta^{(\ell-1)}\phi(\beta^{(\ell-1)}s)\diff s\right)^{T_{\ell-1}-1}
  \\
  &=
  T_{\ell-1}\beta^{(\ell-1)}\phi(\beta^{(\ell-1)}x)
  \left(\int_{-\infty}^{\beta^{(\ell-1)}x}\phi(y)\diff y\right)^{T_{\ell-1}-1}
  \\
  &=
  T_{\ell-1}\beta^{(\ell-1)}\phi(\beta^{(\ell-1)}x)\Phi(\beta^{(\ell-1)}x)^{T_{\ell-1}-1},  
\end{tsaligned}
where the equality in the second line is obtained using
$\cN(x;\mu,\sigma^2)=\frac{1}{\sigma}\phi\left(\frac{x-\mu}{\sigma}\right)$; 
a variable transformation $y=\beta^{(\ell-1)}s$ has been applied
to obtain the third line.
Hence, the second moment of
$z^{(\ell-1)}_{s}$ can be expressed as 
\begin{tsaligned}
  \Ex{\left(z^{(\ell-1)}_s\right)^2}
  &=\Ex{\left(\max\{0,\hat{u}^{(\ell-1)}_s\}\right)^2}
  \Ex[x\sim\hat{p}_{\ell-1}(x)]{\left(\max\{0,x\}\right)^2}
  \\
  &=\int_{-\infty}^\infty\left(\max\{0,x\}\right)^2\hat{p}_{\ell-1}(x)\diff x
  =\int_0^\infty x^2\hat{p}_{\ell-1}(x)\diff x
  \\
  &=T_{\ell-1}\int_0^\infty x^2\beta^{(\ell-1)}
    \phi(\beta^{(\ell-1)}x)\Phi(\beta^{(\ell-1)}x)^{T_{\ell-1}-1}\diff x\\
  &=T_{\ell-1}\int_0^\infty y^2\phi(y)\Phi(y)^{T_{\ell-1}-1}\diff y
  =q^{(\ell-1)}T_{\ell-1}\int_0^\infty y^2\phi(y)\Phi(y)^{T_{\ell-1}-1}\diff y 
  \label{eq:variance-forward-part-maxpool}
\end{tsaligned}
where we have again applied a variable transformation
$y=\beta^{(\ell-1)}x$ to obtain the first equality in the last line. 

\subsubsection{Case of the average pooling}
Recall that $u^{(\ell-1)}_t\sim\cN(0,q^{(\ell-1)})$.
The expectation of each component of 
\begin{tsaligned}
  z^{(\ell-1)}_s=\frac{1}{T_{\ell-1}}\sum_{t\in\vt(\ell,s)}
    \max\left\{0,u^{(\ell-1)}_t\right\}
\end{tsaligned}
can be written as 
\begin{tsaligned}\label{eq:ave-relu-deriv-q}
  \Ex{\max\left\{0,u^{(\ell-1)}_t\right\}}
  &=\int_{-\infty}^\infty
    \max\left\{0,s\right\}\cN(s;0,q^{(\ell-1)})\diff s
    =\int_0^\infty s\cN(s;0,q^{(\ell-1)})\diff s
    \\
  &=\int_0^\infty s\beta^{(\ell-1)}\phi(\beta^{(\ell-1)}s)\diff s
    =\sqrt{q^{(\ell-1)}}\int_0^\infty y\phi(y)\diff y
    =\sqrt{\frac{q^{(\ell-1)}}{2\pi}},  
\end{tsaligned}
where the second equality in the second line follows from a variable transformation $y=\beta^{(\ell-1)}s$.
We again use the variable transformation to rearrange 
the second moment of the component as
\begin{tsaligned}\label{eq:secmom-relu-deriv-q}
  \Ex{\left(\max\left\{0,u^{(\ell-1)}_t\right\}\right)^2}
  &=\int_{-\infty}^\infty\left(\max\{0,s\}\right)^2\cN(s;0,q^{(\ell-1)})\diff s
  =\int_0^\infty s^2\cN(s;0,q^{(\ell-1)})\diff s
  \\
  &=\int_0^\infty s^2\beta^{(\ell-1)}\phi(\beta^{(\ell-1)}s)\diff s\\
  =q^{(\ell-1)}\int_0^\infty y^2\phi(y)\diff y
  =\frac{q^{(\ell-1)}}{2}. 
\end{tsaligned}
The above observations yields
\begin{tsaligned}
  \label{eq:variance-forward-part-avgpool}
  \Ex{\left(z^{(\ell-1)}_s\right)^2}
  &=\frac{1}{T_{\ell-1}^2}\sum_{t\in\vt(\ell,s)}
    \Ex{\left(\max\{0,u^{(\ell-1)}_t\}\right)^2}
    +\frac{1}{T_{\ell-1}^2}\sum_{\substack{
    t,t^\prime\in\vt(\ell,s),\\t\neq t^\prime}}
    \Ex{\max\{0,u^{(\ell-1)}_t\}}\Ex{\max\{0,u^{(\ell-1)}_{t^\prime}\}}
    \\
  &=\frac{1}{T_{\ell-1}}\frac{q^{(\ell-1)}}{2}
    +\frac{T_{\ell-1}-1}{T_{\ell-1}}\frac{q^{(\ell-1)}}{2\pi}
  =q^{(\ell-1)}\left\{\frac{1}{2T_{\ell-1}}
    \left(1+\frac{T_{\ell-1}-1}{\pi}\right)\right\}. 
\end{tsaligned}
\subsection{Derivation of backward variance propagation~\eqref{eq:var-backward}}
\label{sec:appx-backward-variance}
Back-propagation of signal variances, \eqref{eq:var-backward},
can be derived in a similar fashion.
We assume that, in $\ell$th layer ($1<\ell\le L$), each entry in the backward signals $\Delta\z^{(\ell)}$ is iid from $\cN(0,r^{(\ell)})$.
For all $j\in\vj(\ell,j)$, it holds that
\begin{tsaligned}
  \Ex{\left(\Delta z^{(\ell)}_j\right)^2}
  =\Ex{\left(\Delta z^{(\ell)}_j\right)^2}
    -\left(\Ex{\Delta z^{(\ell)}_j}\right)^2=\Var{z^{(\ell)}_j}=r^{(\ell)}.
\end{tsaligned}
Assume that
$\Delta z^{(\ell)}_{d(\ell,j)}$
and
$\partial z^{(\ell)}_{d(\ell,j)}/\partial u^{(\ell)}_j$
are statistically independent to each other.
The expected backward signals $\Delta z^{(\ell-1)}_{i}$ vanishes
as
\begin{tsaligned}
  \Ex{\Delta z^{(\ell-1)}_i}
  &=\Ex{\ip{\vwtil^{(\ell)}_{\tilde{c}(\ell,i),\vh(\ell,i)},
    \Delta\vu^{(\ell)}_{\vj(\ell,i)}}}
  =\sum_{(h,j)\in(\vh(\ell,i),\vj(\ell,i))}
    \Ex{\tilde{w}^{(\ell)}_{\tilde{c}(\ell,i),h}}\Ex{\Delta u^{(\ell)}_j}
  =0. 
\end{tsaligned}
This leads to a recursive expression as
\begin{tsaligned}
  r^{(\ell-1)}_i
  &=\Ex{\left(\Delta z^{(\ell-1)}_i\right)^2}
    -\Ex{\Delta z^{(\ell-1)}_i}^2
  =\Ex{\ip{\vwtil^{(\ell)}_{\tilde{c}(\ell,i),\vh(\ell,i)},
    \Delta\vu^{(\ell)}_{\vj(\ell,i)}}^2}\\
  &=\sum_{(h,j)\in(\vh(\ell,i),\vj(\ell,i))}
    \Ex{\left(\tilde{w}^{(\ell)}_{\tilde{c}(\ell,i),h}\right)^2}
    \Ex{\left(\Delta u^{(\ell)}_j\right)^2}
  +\sum_{\substack{(h,j),(h^\prime,j^\prime)
    \in(\vh(\ell,i),\vj(\ell,i)),\\h\neq h^\prime,j\neq j^\prime}}
    \hspace{-3.5em}
    \Ex{\tilde{w}^{(\ell)}_{\tilde{c}(\ell,i),h}}
    \Ex{\tilde{w}^{(\ell)}_{\tilde{c}(\ell,i),h^\prime}}
    \Ex{\Delta u^{(\ell)}_j\Delta u^{(\ell)}_{j^\prime}}\\
  &=\sigma^2_{w^{(\ell)}}\sum_{j\in\vj(\ell,i)}
    \Ex{\left(\Delta u^{(\ell)}_j\right)^2}
  =\sigma^2_{w^{(\ell)}}\sum_{j\in\vj(\ell,i)}
    \Ex{\left(\Delta z^{(\ell)}_{d(\ell,j)}\right)^2}
    \Ex{\left(\pderiv{z^{(\ell)}_{d(\ell,j)}}{u^{(\ell)}_j}\right)^2}\\
  &=\sigma^2_{w^{(\ell)}}r^{(\ell)}\sum_{j\in\vj(\ell,i)}
    \Ex{\left(\pderiv{z^{(\ell)}_{d(\ell,j)}}{u^{(\ell)}_j}\right)^2}. 
  \label{eq:variance-backward-base}
\end{tsaligned}
The gradient $\pderiv{z^{(\ell)}_{d(\ell,j)}}{u^{(\ell)}_j}$
depends on the types of the activation and the pooling operations.
We shall derive
\begin{tsaligned}\label{eq:secmom-z-deriv-backward}
  \Ex{\left(\pderiv{z^{(\ell)}_{d(\ell,j)}}{u^{(\ell)}_j}\right)^2}
  =
  \gamma_\ell J_{\ell,i} 
\end{tsaligned}
where $J_{\ell,i}=|\vj(\ell,i)|$, for two cases: 
ReLU activation followed by the max pooling and
ReLU activation followed by the average pooling.
The recursive expression \eqref{eq:var-backward}
will then be derived by substituting
\eqref{eq:secmom-z-deriv-backward} into
\eqref{eq:variance-backward-base}, and
take an approximation. 

\subsubsection{Case for the max pooling}
Function $\hat{u}^{(\ell)}_{i}\mapsto z^{(\ell)}_i=\max\{0,\hat{u}^{(\ell)}_{i}\}$
is not differentiable at $\hat{u}^{(\ell)}_{i}=0$.
A subgradient at $\hat{u}^{(\ell)}_{i}=0$ is one, yielding
\begin{tsaligned}\label{eq:def-U-deriv-backward}
  \pderiv{z^{(\ell)}_{d(\ell,j)}}{\hat{u}^{(\ell)}_{d(\ell,j)}}
  = U(\hat{u}^{(\ell)}_{d(\ell,j)}),
  \quad\text{ where }\quad
  U(x)
  :=
  \begin{cases}
    1&\text{if}~x \ge  0,\\
    0&\text{otherwise}. 
  \end{cases}
\end{tsaligned}
Using
\begin{tsaligned}
  \pderiv{\hat{u}^{(\ell)}_{d(\ell,j)}}{u^{(\ell)}_j}
  &=\delta_{\hat{u}^{(\ell)}_{d(\ell,j)},u^{(\ell)}_j}
\end{tsaligned}
we have 
\begin{tsaligned}
  \pderiv{z^{(\ell)}_{d(\ell,j)}}{u^{(\ell)}_j}
  =\pderiv{z^{(\ell)}_{d(\ell,j)}}{\hat{u}^{(\ell)}_{d(\ell,j)}}
    \pderiv{\hat{u}^{(\ell)}_{d(\ell,j)}}{u^{(\ell)}_j}
    =
    \begin{cases}
    1&\text{if}~\hat{u}^{(\ell)}_{d(\ell,j)}\geq0~\text{and}~
    \hat{u}^{(\ell)}_{d(\ell,j)}=u^{(\ell)}_j,\\
    0&\text{otherwise}. 
  \end{cases}
\end{tsaligned}
Under the iid assumption of $u^{(\ell)}_{i}$, 
it holds that 
\begin{tsaligned}
  \Pr\left(\hat{u}^{(\ell)}_{d(\ell,j)}=u^{(\ell)}_j\right)
  &=\frac{1}{T_\ell},
\end{tsaligned}
and 
\begin{tsaligned}
  \Pr\left(\hat{u}^{(\ell)}_{d(\ell,j)}\geq0\right)
  =1-\Pr\left(\hat{u}^{(\ell)}_{d(\ell,j)}<0\right)
  =1-\prod_{t\in\vt(\ell,d(\ell,j))}\Pr\left(u^{(\ell)}_t<0\right)
  =1-\Phi(0)^{T_\ell}
  =1-2^{-T_\ell}. 
\end{tsaligned}
Hence, we obtain
\begin{tsaligned}
  \Ex{\left(\pderiv{z^{(\ell)}_{d(\ell,j)}}{u^{(\ell)}_j}\right)^2}
  =\Pr(\hat{u}^{(\ell)}_{d(\ell,j)}\geq0)
    \Pr(\hat{u}^{(\ell)}_{d(\ell,j)}=u^{(\ell)}_j)
  =\frac{1-2^{-T_\ell}}{T_\ell}
  =\frac{2^{T_\ell}-1}{T_\ell2^{T_\ell}}
  \label{eq:variance-backward-part-maxpool}
\end{tsaligned}
resulting in \eqref{eq:secmom-z-deriv-backward}. 

\subsubsection{Case for the average pooling}
The gradient of the average pooling is expressed as
\begin{tsaligned}
  \pderiv{z^{(\ell)}_{d(\ell,j)}}{v^{(\ell)}_j}=\frac{1}{T_\ell}. 
\end{tsaligned}
We reuse the unit step function defined in \eqref{eq:def-U-deriv-backward}
to get
\begin{tsaligned}
  \label{eq:variance-backward-part-avgpool}
  \Ex{\left(\pderiv{z^{(\ell)}_{d(\ell,j)}}{u^{(\ell)}_j}\right)^2}
  &=\Ex{\left(\pderiv{z^{(\ell)}_{d(\ell,j)}}{v^{(\ell)}_j}
    \pderiv{v^{(\ell)}_j}{u^{(\ell)}_j}\right)^2}
  =\frac{1}{T_\ell^2}\Ex{U\left(u^{(\ell)}_j\right)^2}\\
  &=\frac{1}{T_\ell^2}\cdot1\cdot\Pr\left(u^{(\ell)}_j\geq0\right)
  =\frac{1}{T_\ell^2}\left(1-\Pr\left(u^{(\ell)}_j<0\right)\right)
  =\frac{1-\Phi(0)}{T_\ell^2}
  =\frac{1}{2T_\ell^2}. 
\end{tsaligned}
yielding \eqref{eq:secmom-z-deriv-backward}.

\subsubsection{Approximation of variances}
The variance of the backward signals at $i$th unit,
say $r^{(\ell-1)}_i$, may not be equal to those of some units. 
If we approximate $r^{(\ell-1)}_i$ with
the average over units, we have 
\begin{tsaligned}
  \label{eq:variance-backward-deriv-backward}
  r^{(\ell-1)}
  =\frac{1}{M_{\ell-1}}\sum_{i=1}^{M_{\ell-1}}r^{(\ell-1)}_i
  =\sigma^2_{w^{(\ell)}}r^{(\ell)}\gamma_{\ell}
    \frac{1}{M_{\ell-1}}\varepsilon_\ell.
\end{tsaligned}
where $\varepsilon_\ell$ is the total number of connections
between $(\ell-1)$th and $\ell$th layers, namely, 
$\varepsilon_\ell=\sum_{i=1}^{M_{\ell-1}}J_{\ell,i}$. 
Thus, the recursive expression \eqref{eq:var-backward} has been derived.

\section{Additional Experimental Results}
\label{sec:addexp}
First, we report how we have determined the experimental settings in which the experiments described in the main text were carried out. We conducted a preliminary experiment for choosing an optimizer from SGD, Adadelta, Adagrad, RMSprop, and Adam. The batch size was set to 64 that was the maximal size of our computational environment. Eight learning rates $10^{0},10^{-1},\dots, 10^{-7}$ were used. We measured the validation accuracies at 50th epoch on the dataset Car. Table~\ref{tab:pre}a shows the maximum accuracies among those of eight learning rates. The highest validation accuracy among all initialization methods and all optimizers was obtained with ASV backward and Adam. When we look at the accuracies in each row of Table~\ref{tab:pre}a, it is observed that one of the two ASV methods performed better than Kaiming and Xavier for every optimizer. Comparison of the accuracies in each column of Table~\ref{tab:pre} yields a suggestion that Adam is the best optimizer for any of five initialization methods, which made us to opt Adam as the optimizer for our main simulations. 

We next see the validation accuracies obtained with different learning rates to determine the range of the learning rate used for our main experiments. The performances reported in Table~\ref{tab:pre}b suggest that although too small or too large learning rates made learning stagnant, an effective learning rate is between $10^{-3}$ and $10^{-6}$ for all initialization methods. 

We examined the proposed initialization methods on a deeper convolutional neural architecture. In addition to the 34-layer architecture described in Table~\ref{tab:arch}, we used a 50-layer architecture detailed in Table~\ref{tab:arch-f50}. The experimental settings are common to those discussed in the main text, except the architecture used for the simulation. The pattern recognition performances are reported in Table~\ref{tab:acc-f50}. Similarly to Table~\ref{tab:acc.g2}, ASV backward method again achieved the best performance among all the other initialization methods. If we compare Table~\ref{tab:arch-f50} with Table~\ref{tab:arch}, it can also be observed that the accuracies of 50 layers are slightly worse than the accuracies of 34 layers. These imply that the proposed methods work well even with deeper neural networks, although some heuristics such as data augmentation might be introduced together to obtain a better pattern recognition accuracy.

We have seen the validation accuracies at 1,000th epoch in Table~\ref{tab:evolution-f34-car}, whereas the accuracies at earlier epochs are reported in Table~\ref{tab:evolution-f34-car}. For any initialization method, the improvement of the validation accuracies was decelerated after exceeding 700th epoch. We observe that no method could improve the performance by 1\% within 100 epochs after 700th epoch, making it hard to expect that better accuracies is obtained by taking more epochs. From these results, we have decided that 1,000th epoch is used for learning in the main experiments.
\begin{table*}[t!]
  \centering
  \caption{Structure of 50-layer architecture. The definition of the \texttt{InputBlock} is in Table~\ref{tab:arch}. In addition to the 34-layer architecture, this model includes a structure in which the number of channels decreases once in the intermediate layer. It was verified that the proposed method worked effectively even with this structure.
  \label{tab:arch-f50}}
  \begin{tabular}{|l|}
    \hline
    Component \verb|ConvBlock2|($c_1,c_2,s=1$)\\
    \hline
    Conv(1x1,channels=$c_1$,padding=0,stride=1),\\
    Conv(3x3,channels=$c_1$,padding=1,stride=$s$),\\
    Conv(1x1,channels=$c_2$,padding=0,stride=1)\\
    \hline
  \end{tabular}\\\vspace{1ex}
  \begin{tabular}{|c|c|l|}
    \hline
    $\ell$-th Layer&Output Shape&Model \verb|F50|\\
    \hline
    \hline
    &(3,224,224) &Input Image\\
    \hline
    1&(64,112,112)&\verb|InputBlock|($c=64$)\\
    \hline
    $2$--$ 10$&(256,56,56)&\verb|ConvBlock2|($c_1=64,c_2=256$)$\times3$\\
    \hline
    $11$--$ 22$ &(512,28,28)&\verb|ConvBlock2|($c_1=128,c_2=512,s=2$)\\
    &&\verb|ConvBlock2|($c_1=128,c_2=512$)$\times3$\\
    \hline
    $23$--$ 40$ & (1024,14,14) &\verb|ConvBlock2|($c_1=256,c_2=1024,s=2$)\\
    &&\verb|ConvBlock2|($c_1=256,c_2=1024$)$\times5$\\
    \hline
    $41$--$ 49$ & (2048,7,7) &\verb|ConvBlock2|($c_1=512,c_2=2048,s=2$)\\
    &&\verb|ConvBlock2|($c_1=512,c_2=2048$)$\times2$\\
    %\hline
    &(2048,1,1)   &Global Average Pooling\\
    \hline
    $50$&10         &Linear\\
    \hline
    \hline
    &$2.07\times10^7$&Number of Parameters\\
    \hline
  \end{tabular}
\end{table*}

\begin{table*}[t!]
  \centering
  \caption{
  	%Validation accuracies of the 50-layer architecture. 
  	Experimental results for the dataset Car using 50-layer architecture. The notation used is the same as in Table~\ref{tab:acc.g2}. Similar to Table~\ref{tab:acc.g2}a, the proposed backward method yields the highest performance but it is slightly lower than that of the obtained performance using 34-layer architecture. Due to this and considering the increase in computational cost, there is no advantage of adopting 50-layer architecture for 34-layer architecture.
    \label{tab:acc-f50}}
  %% \label{tbl:results}
  \begin{tabular}{l}
    % (a) Car
    % \\
    \begin{tabular}{|c|c|c|c|c|c|}
      \hline
      50-layer &
      \multicolumn{5}{|c|}{Initialization Methods} \\
      \hline
      Learning & \multirow{2}{*}{Xavier}  &Kaiming  &Kaiming   & ASV & ASV \\
      Rate $(lr)$    &         &(forward)&(backward)&(forward)&(backward)\\
    \hline
    \hline
    $10^{-3}$&15.78&15.78&15.78&15.78&15.78\\
    $10^{-4}$&15.78&15.78&15.78&15.78&\textit{\textbf{80.77}}\\
    $10^{-5}$&\textit{69.23}&\textit{69.87}&\textit{71.23}&\textit{69.58}&66.43\\
    $10^{-6}$&33.07&55.45&54.95&53.95&53.16\\
    \hline
    \end{tabular}
  \end{tabular}
\end{table*}
\begin{table}
  % taken from sheet 'compare-optim', 2020/08/13
  % verify: OK (2020/08/13)
  %  (a) ok
  %  (b) ok
  \caption{Experimental results in varying learning conditions
    \label{tab:pre}}
  \centering
  \begin{tabular}{p{\linewidth}}
    (a) Experimental results comparing the effects of initialization methods on different optimization algorithms. For each initialization method and optimizer, the learning rate $lr$ was set from $lr = 10^{0}$ to $10^{-7}$, with a step factor of $1/10$. The values indicated show the maximum accuracy rate for the testing data at the particular $lr$. Of all the optimizers, Adam achieved the highest recognition performance for all the initialization methods and through this, Adam was implemented as the optimization algorithm in the main experiments. In particular, it can be pointed out that the proposed initialization method performed best among all the optimization methods.
    \\
    \center{
    \begin{tabular}{|c|c|c|c|c|c|}
      \hline
      \multirow{3}{*}{Optimizer}  &
      \multicolumn{5}{c|}{Initialization Methods (Max Accuracy/$lr$)} \\
      \cline{2-6}
       & \multirow{2}{*}{Xavier}  &Kaiming  &Kaiming   & ASV & ASV \\
       &         &(forward)&(backward)&(forward)&(backward)\\
      \hline
      \hline
      SGD&15.78/all&35.15/$10^{-2}$ & 31.78/$10^{-2}$ & 35.29/$10^{-2}$ & 24.39/$10^{-3}$ \\
      \hline
      Adadelta&15.78/all&38.52/$10^{-1}$ & 39.53/$10^{-2}$ & 42.04/$10^{-1}$ & 50.22/$10^{-1}$ \\
      \hline
      Adagrad&27.19/$10^{-4}$ & 36.37/$10^{-4}$ & 34.79/$10^{-4}$ & 43.33/$10^{-3}$ & 35.58/$10^{-3}$ \\
      \hline
      RMSprop&28.98/$10^{-5}$ & 41.75/$10^{-5}$ & 42.40/$10^{-5}$ & 42.47/$10^{-5}$ & 35.80/$10^{-5}$ \\
      \hline
      Adam&31.56/$10^{-5}$ & 45.05/$10^{-5}$ & 48.06/$10^{-5}$ & 44.76/$10^{-5}$ & \textbf{51.15}/$10^{-4}$ \\
      \hline
    \end{tabular}
	}
    \\
    \flushleft{
    (b) Experimental results in investigating the effect of the learning rates when using Adam as the optimization algorithm. Even if Adam, which is adaptive algorithm, is used, the learning rate $lr$ needs to be still adjusted. Learning becomes stagnant when $lr$ is too large or too small (which is observed for performance values such as $8.68$ and $15.78$). In the main experiments, we decided to limit the learning rate to $lr = 10^{-3}$ to $10^{-6}$ so that the learning will be possible.}
    \\
    \centering{
    \begin{tabular}{|c||c|c|c|c|c|}
      \hline
      & \multicolumn{5}{c|}{Initialization Methods} \\
      \cline{2-6}
      Learning & \multirow{2}{*}{Xavier}  &Kaiming  &Kaiming   & ASV & ASV \\
      Rate $(lr)$     &         &(forward)&(backward)&(forward)&(backward)\\
      \hline
      \hline
      $10^{-0}$& 8.68& 8.68& 8.68& 8.68& 8.68\\
      \hline
      $10^{-1}$& 15.78& 8.68& 8.68& 8.68& 8.68\\
      \hline
      $10^{-2}$& 15.78& 15.78& 15.78& 15.78& 15.78\\
      \hline
      $10^{-3}$& 15.78& 15.78& 15.78& 15.78& 15.78\\
      \hline
      $10^{-4}$& 15.78& 15.78& 15.78& 15.78&\textbf{51.15}\\
      \hline
      $10^{-5}$&31.56&45.05&48.06&44.76&33.21\\
      \hline
      $10^{-6}$& 15.78&27.04&25.90&27.76&23.67\\
      \hline
      $10^{-7}$& 15.78& 15.78& 15.78& 15.78& 16.21\\
      \hline
    \end{tabular}
}
  \end{tabular}
\end{table}
\begin{table}
  % taken from $results/g2, peak condition, 2020/08/13
  % verify: OK (2020/08/13)
  %  xa(lr=1e-5) ok
  %  ka_f(lr=1e-5) ok
  %  ka_b(lr=1e-5) ok
  %  cn_f(lr=1e-5) ok
  %  cn_b(lr=1e-4) ok
  \caption{
%  	Validation accuracies agaist number of epochs.
%    Numbers in brackets indicate the improvement
%    from the validation accuracy at 100 epochs before the iteration.
	Performance evolution of learning for Table~\ref{tab:acc.g2}a. The numerical value is the validation accuracy and the number in parentheses is the difference in current accuracy rate from the previous column (100 epochs ago). 
  \label{tab:evolution-f34-car}}
  \centering
  \def\xg#1{{(\scriptsize+#1)}}
  \begin{tabular}{|c||c|c|c|c|c|c|c|c|c|c|}
    \hline
    Initialization & \multicolumn{10}{c|}{Epoch} \\
    \cline{2-11}
    Methods&100&200&300&400&500&600&700&800&900&1000\\
    \hline
    \hline
    \multirow{2}{*}{Xavier}&42.11&56.38&61.84&64.92&66.43&68.58&69.87&71.31&71.31&71.95\\
    &&\xg{14.28}&\xg{5.45}&\xg{3.08}&\xg{1.51}&\xg{2.15}&\xg{1.29}&\xg{1.43}&\xg{0.00}&\xg{0.65}\\
    \hline
    Kaiming&54.16&63.92&66.50&68.44&68.87&68.94&69.66&70.09&70.52&70.52\\
    (forward)&&\xg{9.76}&\xg{2.58}&\xg{1.94}&\xg{0.43}&\xg{0.07}&\xg{0.72}&\xg{0.43}&\xg{0.43}&\xg{0.00}\\
    \hline
    Kaiming&55.81&63.20&68.01&69.37&69.87&71.31&71.88&73.10&73.10&73.10\\
    (backward)&&\xg{7.39}&\xg{4.81}&\xg{1.36}&\xg{0.50}&\xg{1.43}&\xg{0.57}&\xg{1.22}&\xg{0.00}&\xg{0.00}\\
    \hline
    Proposed&56.46&63.06&66.86&68.01&69.58&70.80&71.23&71.59&72.74&72.74\\
    (forward)&&\xg{6.60}&\xg{3.80}&\xg{1.15}&\xg{1.58}&\xg{1.22}&\xg{0.43}&\xg{0.36}&\xg{1.15}&\xg{0.00}\\
    \hline
    Proposed&67.50&75.75&79.41&80.20&81.06&81.06&81.21&81.42&81.49&81.49\\
    (backward)&&\xg{8.25}&\xg{3.66}&\xg{0.79}&\xg{0.86}&\xg{0.00}&\xg{0.14}&\xg{0.22}&\xg{0.07}&\xg{0.00}\\
    \hline
  \end{tabular}
\end{table}

\end{document}